\documentclass{article}

\usepackage[preprint]{neurips_2026}

\usepackage[utf8]{inputenc}
\usepackage[T1]{fontenc}
\usepackage{hyperref}
\usepackage{url}
\usepackage{booktabs}
\usepackage{amsmath}
\usepackage{amssymb}
\usepackage{microtype}
\usepackage{xcolor}
\usepackage{graphicx}
\usepackage{array}
\usepackage{multirow}
\usepackage{enumitem}
\usepackage{listings}
\usepackage{pifont}
\usepackage{colortbl}
\usepackage{framed}

\usepackage{standalone}
\usepackage{tikz}
\usetikzlibrary{arrows.meta,decorations.pathreplacing,shapes.symbols}

\definecolor{segsys}{HTML}{F4E4DA}
\definecolor{segtool}{HTML}{C1D8D2}
\definecolor{segprob}{HTML}{E8CBB9}
\definecolor{segthink}{HTML}{95BEB6}
\definecolor{segnudge}{HTML}{DAA88D}
\definecolor{segclose}{HTML}{D6E5E0}
\definecolor{figfg}{HTML}{232019}
\definecolor{figfg2}{HTML}{6B6459}
\definecolor{figline}{HTML}{E7E1D5}
\definecolor{cutred}{HTML}{B4531F}
\definecolor{figteal}{HTML}{0F6D5F}
\definecolor{rolepre}{HTML}{1F6674}  \definecolor{roleint}{HTML}{904B72}  \definecolor{rolegen}{HTML}{60722F}  \definecolor{rolebase}{HTML}{56606B}

\newcolumntype{S}{>{\raggedright\arraybackslash}p{2.42cm}}
\newcolumntype{H}{>{\raggedright\arraybackslash}p{3.05cm}}

\newcolumntype{T}{@{}>{\ttfamily\scriptsize}p{13.10cm}@{}}

\newcommand{\tool}{\textsc{tool}}
\newcommand{\self}{\textsc{self}}
\newcommand{\other}{\textsc{other}}
\newcommand{\selftotool}{\self\,$\to$\,\tool}
\newcommand{\tooltoself}{\tool\,$\to$\,\self}
\newcommand{\ind}{\mathbb{I}}

\newcommand{\pp}{\,\mathrm{pp}}

\newcommand{\nudgepos}{\nu^{+}}
\newcommand{\nudgeneg}{\nu^{-}}
\newcommand{\nudgeneutral}{\nu^{0}}
\newcommand{\none}{\varnothing}

\newif\ifshowcrosstransplants
\showcrosstransplantsfalse
\definecolor{crossTransplantBg}{gray}{0.94}

\newenvironment{crossTransplantBlock}{\MakeFramed{\advance\hsize-\width\FrameRestore}\noindent\ignorespaces
}{\endMakeFramed}

\hypersetup{
  colorlinks=true,
  linkcolor=blue,
  citecolor=blue,
  urlcolor=blue,
  pdftitle={Nudgeability: Reasoning Models Follow Confidence Signals Without Tracking Their Own Competence},
  pdfauthor={}
}
\hypersetup{pdfauthor={Rohit Saxena; Utkarsh Upadhyay}}

\title{Nudgeability: Reasoning Models Follow Confidence\\Signals Without Tracking Their Own Competence}

\author{Rohit Saxena
  \And
  Utkarsh Upadhyay
}

\providecommand{\newEvidence}[3]{\csname newEvidence@#1@#2@#3\endcsname}
\expandafter\def\csname newEvidence@musique_2hop@qwen3-4b-fp8@S-point\endcsname{14.0}
\expandafter\def\csname newEvidence@musique_2hop@qwen3-4b-fp8@S-low\endcsname{11.8}
\expandafter\def\csname newEvidence@musique_2hop@qwen3-4b-fp8@S-high\endcsname{16.3}
\expandafter\def\csname newEvidence@musique_2hop@qwen3-4b-fp8@S-ci\endcsname{14.0 [11.8, 16.3]}
\expandafter\def\csname newEvidence@musique_2hop@qwen3-4b-fp8@wt-point\endcsname{58.2}
\expandafter\def\csname newEvidence@musique_2hop@qwen3-4b-fp8@wt-low\endcsname{49.2}
\expandafter\def\csname newEvidence@musique_2hop@qwen3-4b-fp8@wt-high\endcsname{67.2}
\expandafter\def\csname newEvidence@musique_2hop@qwen3-4b-fp8@wt-ci\endcsname{58.2 [49.2, 67.2]}
\expandafter\def\csname newEvidence@musique_2hop@qwen3-4b-fp8@null-point\endcsname{39.4}
\expandafter\def\csname newEvidence@musique_2hop@qwen3-4b-fp8@null-low\endcsname{35.7}
\expandafter\def\csname newEvidence@musique_2hop@qwen3-4b-fp8@null-high\endcsname{43.1}
\expandafter\def\csname newEvidence@musique_2hop@qwen3-4b-fp8@null-ci\endcsname{39.4 [35.7, 43.1]}
\expandafter\def\csname newEvidence@musique_2hop@qwen3-4b-fp8@lift-point\endcsname{18.8}
\expandafter\def\csname newEvidence@musique_2hop@qwen3-4b-fp8@lift-low\endcsname{10.7}
\expandafter\def\csname newEvidence@musique_2hop@qwen3-4b-fp8@lift-high\endcsname{27.1}
\expandafter\def\csname newEvidence@musique_2hop@qwen3-4b-fp8@lift-ci\endcsname{18.8 [10.7, 27.1]}
\expandafter\def\csname newEvidence@musique_2hop@qwen3-4b-fp8@n_S\endcsname{966}
\expandafter\def\csname newEvidence@musique_2hop@qwen3-4b-fp8@n_wt\endcsname{876}
\expandafter\def\csname newEvidence@musique_2hop@qwen3-4b-fp8@n_binary_eligible_union\endcsname{876}
\expandafter\def\csname newEvidence@musique_2hop@qwen3-4b-fp8@flips\endcsname{134}
\expandafter\def\csname newEvidence@musique_2hop@qwen3-4b-fp8@helpful_flips\endcsname{78}
\expandafter\def\csname newEvidence@musique_2hop@qwen3-8b-fp8@S-point\endcsname{22.4}
\expandafter\def\csname newEvidence@musique_2hop@qwen3-8b-fp8@S-low\endcsname{19.6}
\expandafter\def\csname newEvidence@musique_2hop@qwen3-8b-fp8@S-high\endcsname{25.1}
\expandafter\def\csname newEvidence@musique_2hop@qwen3-8b-fp8@S-ci\endcsname{22.4 [19.6, 25.1]}
\expandafter\def\csname newEvidence@musique_2hop@qwen3-8b-fp8@wt-point\endcsname{42.5}
\expandafter\def\csname newEvidence@musique_2hop@qwen3-8b-fp8@wt-low\endcsname{36.4}
\expandafter\def\csname newEvidence@musique_2hop@qwen3-8b-fp8@wt-high\endcsname{48.8}
\expandafter\def\csname newEvidence@musique_2hop@qwen3-8b-fp8@wt-ci\endcsname{42.5 [36.4, 48.8]}
\expandafter\def\csname newEvidence@musique_2hop@qwen3-8b-fp8@null-point\endcsname{40.6}
\expandafter\def\csname newEvidence@musique_2hop@qwen3-8b-fp8@null-low\endcsname{37.1}
\expandafter\def\csname newEvidence@musique_2hop@qwen3-8b-fp8@null-high\endcsname{44.1}
\expandafter\def\csname newEvidence@musique_2hop@qwen3-8b-fp8@null-ci\endcsname{40.6 [37.1, 44.1]}
\expandafter\def\csname newEvidence@musique_2hop@qwen3-8b-fp8@lift-point\endcsname{1.9}
\expandafter\def\csname newEvidence@musique_2hop@qwen3-8b-fp8@lift-low\endcsname{-3.1}
\expandafter\def\csname newEvidence@musique_2hop@qwen3-8b-fp8@lift-high\endcsname{7.1}
\expandafter\def\csname newEvidence@musique_2hop@qwen3-8b-fp8@lift-ci\endcsname{1.9 [-3.1, 7.1]}
\expandafter\def\csname newEvidence@musique_2hop@qwen3-8b-fp8@n_S\endcsname{944}
\expandafter\def\csname newEvidence@musique_2hop@qwen3-8b-fp8@n_wt\endcsname{921}
\expandafter\def\csname newEvidence@musique_2hop@qwen3-8b-fp8@n_binary_eligible_union\endcsname{918}
\expandafter\def\csname newEvidence@musique_2hop@qwen3-8b-fp8@flips\endcsname{301}
\expandafter\def\csname newEvidence@musique_2hop@qwen3-8b-fp8@helpful_flips\endcsname{128}
\expandafter\def\csname newEvidence@musique_2hop@qwen3-14b-fp8@S-point\endcsname{31.0}
\expandafter\def\csname newEvidence@musique_2hop@qwen3-14b-fp8@S-low\endcsname{28.0}
\expandafter\def\csname newEvidence@musique_2hop@qwen3-14b-fp8@S-high\endcsname{34.0}
\expandafter\def\csname newEvidence@musique_2hop@qwen3-14b-fp8@S-ci\endcsname{31.0 [28.0, 34.0]}
\expandafter\def\csname newEvidence@musique_2hop@qwen3-14b-fp8@wt-point\endcsname{41.7}
\expandafter\def\csname newEvidence@musique_2hop@qwen3-14b-fp8@wt-low\endcsname{36.4}
\expandafter\def\csname newEvidence@musique_2hop@qwen3-14b-fp8@wt-high\endcsname{47.1}
\expandafter\def\csname newEvidence@musique_2hop@qwen3-14b-fp8@wt-ci\endcsname{41.7 [36.4, 47.1]}
\expandafter\def\csname newEvidence@musique_2hop@qwen3-14b-fp8@null-point\endcsname{36.5}
\expandafter\def\csname newEvidence@musique_2hop@qwen3-14b-fp8@null-low\endcsname{33.5}
\expandafter\def\csname newEvidence@musique_2hop@qwen3-14b-fp8@null-high\endcsname{39.6}
\expandafter\def\csname newEvidence@musique_2hop@qwen3-14b-fp8@null-ci\endcsname{36.5 [33.5, 39.6]}
\expandafter\def\csname newEvidence@musique_2hop@qwen3-14b-fp8@lift-point\endcsname{5.2}
\expandafter\def\csname newEvidence@musique_2hop@qwen3-14b-fp8@lift-low\endcsname{0.9}
\expandafter\def\csname newEvidence@musique_2hop@qwen3-14b-fp8@lift-high\endcsname{9.6}
\expandafter\def\csname newEvidence@musique_2hop@qwen3-14b-fp8@lift-ci\endcsname{5.2 [0.9, 9.6]}
\expandafter\def\csname newEvidence@musique_2hop@qwen3-14b-fp8@n_S\endcsname{990}
\expandafter\def\csname newEvidence@musique_2hop@qwen3-14b-fp8@n_wt\endcsname{986}
\expandafter\def\csname newEvidence@musique_2hop@qwen3-14b-fp8@n_binary_eligible_union\endcsname{981}
\expandafter\def\csname newEvidence@musique_2hop@qwen3-14b-fp8@flips\endcsname{321}
\expandafter\def\csname newEvidence@musique_2hop@qwen3-14b-fp8@helpful_flips\endcsname{134}
\expandafter\def\csname newEvidence@musique_2hop@qwen3-32b-fp8@S-point\endcsname{38.3}
\expandafter\def\csname newEvidence@musique_2hop@qwen3-32b-fp8@S-low\endcsname{35.1}
\expandafter\def\csname newEvidence@musique_2hop@qwen3-32b-fp8@S-high\endcsname{41.3}
\expandafter\def\csname newEvidence@musique_2hop@qwen3-32b-fp8@S-ci\endcsname{38.3 [35.1, 41.3]}
\expandafter\def\csname newEvidence@musique_2hop@qwen3-32b-fp8@wt-point\endcsname{37.2}
\expandafter\def\csname newEvidence@musique_2hop@qwen3-32b-fp8@wt-low\endcsname{32.1}
\expandafter\def\csname newEvidence@musique_2hop@qwen3-32b-fp8@wt-high\endcsname{42.2}
\expandafter\def\csname newEvidence@musique_2hop@qwen3-32b-fp8@wt-ci\endcsname{37.2 [32.1, 42.2]}
\expandafter\def\csname newEvidence@musique_2hop@qwen3-32b-fp8@null-point\endcsname{30.4}
\expandafter\def\csname newEvidence@musique_2hop@qwen3-32b-fp8@null-low\endcsname{27.4}
\expandafter\def\csname newEvidence@musique_2hop@qwen3-32b-fp8@null-high\endcsname{33.3}
\expandafter\def\csname newEvidence@musique_2hop@qwen3-32b-fp8@null-ci\endcsname{30.4 [27.4, 33.3]}
\expandafter\def\csname newEvidence@musique_2hop@qwen3-32b-fp8@lift-point\endcsname{6.8}
\expandafter\def\csname newEvidence@musique_2hop@qwen3-32b-fp8@lift-low\endcsname{3.0}
\expandafter\def\csname newEvidence@musique_2hop@qwen3-32b-fp8@lift-high\endcsname{10.7}
\expandafter\def\csname newEvidence@musique_2hop@qwen3-32b-fp8@lift-ci\endcsname{6.8 [3.0, 10.7]}
\expandafter\def\csname newEvidence@musique_2hop@qwen3-32b-fp8@n_S\endcsname{975}
\expandafter\def\csname newEvidence@musique_2hop@qwen3-32b-fp8@n_wt\endcsname{981}
\expandafter\def\csname newEvidence@musique_2hop@qwen3-32b-fp8@n_binary_eligible_union\endcsname{959}
\expandafter\def\csname newEvidence@musique_2hop@qwen3-32b-fp8@flips\endcsname{366}
\expandafter\def\csname newEvidence@musique_2hop@qwen3-32b-fp8@helpful_flips\endcsname{136}
\expandafter\def\csname newEvidence@musique_2hop@gemma4-e2b-w4a16@S-point\endcsname{24.2}
\expandafter\def\csname newEvidence@musique_2hop@gemma4-e2b-w4a16@S-low\endcsname{21.5}
\expandafter\def\csname newEvidence@musique_2hop@gemma4-e2b-w4a16@S-high\endcsname{27.0}
\expandafter\def\csname newEvidence@musique_2hop@gemma4-e2b-w4a16@S-ci\endcsname{24.2 [21.5, 27.0]}
\expandafter\def\csname newEvidence@musique_2hop@gemma4-e2b-w4a16@wt-point\endcsname{36.9}
\expandafter\def\csname newEvidence@musique_2hop@gemma4-e2b-w4a16@wt-low\endcsname{31.1}
\expandafter\def\csname newEvidence@musique_2hop@gemma4-e2b-w4a16@wt-high\endcsname{42.9}
\expandafter\def\csname newEvidence@musique_2hop@gemma4-e2b-w4a16@wt-ci\endcsname{36.9 [31.1, 42.9]}
\expandafter\def\csname newEvidence@musique_2hop@gemma4-e2b-w4a16@null-point\endcsname{44.7}
\expandafter\def\csname newEvidence@musique_2hop@gemma4-e2b-w4a16@null-low\endcsname{41.1}
\expandafter\def\csname newEvidence@musique_2hop@gemma4-e2b-w4a16@null-high\endcsname{48.3}
\expandafter\def\csname newEvidence@musique_2hop@gemma4-e2b-w4a16@null-ci\endcsname{44.7 [41.1, 48.3]}
\expandafter\def\csname newEvidence@musique_2hop@gemma4-e2b-w4a16@lift-point\endcsname{-7.9}
\expandafter\def\csname newEvidence@musique_2hop@gemma4-e2b-w4a16@lift-low\endcsname{-12.6}
\expandafter\def\csname newEvidence@musique_2hop@gemma4-e2b-w4a16@lift-high\endcsname{-3.0}
\expandafter\def\csname newEvidence@musique_2hop@gemma4-e2b-w4a16@lift-ci\endcsname{-7.9 [-12.6, -3.0]}
\expandafter\def\csname newEvidence@musique_2hop@gemma4-e2b-w4a16@n_S\endcsname{1000}
\expandafter\def\csname newEvidence@musique_2hop@gemma4-e2b-w4a16@n_wt\endcsname{993}
\expandafter\def\csname newEvidence@musique_2hop@gemma4-e2b-w4a16@n_binary_eligible_union\endcsname{993}
\expandafter\def\csname newEvidence@musique_2hop@gemma4-e2b-w4a16@flips\endcsname{293}
\expandafter\def\csname newEvidence@musique_2hop@gemma4-e2b-w4a16@helpful_flips\endcsname{108}
\expandafter\def\csname newEvidence@musique_2hop@gemma4-e4b-w4a16@S-point\endcsname{66.5}
\expandafter\def\csname newEvidence@musique_2hop@gemma4-e4b-w4a16@S-low\endcsname{63.5}
\expandafter\def\csname newEvidence@musique_2hop@gemma4-e4b-w4a16@S-high\endcsname{69.5}
\expandafter\def\csname newEvidence@musique_2hop@gemma4-e4b-w4a16@S-ci\endcsname{66.5 [63.5, 69.5]}
\expandafter\def\csname newEvidence@musique_2hop@gemma4-e4b-w4a16@wt-point\endcsname{29.1}
\expandafter\def\csname newEvidence@musique_2hop@gemma4-e4b-w4a16@wt-low\endcsname{25.6}
\expandafter\def\csname newEvidence@musique_2hop@gemma4-e4b-w4a16@wt-high\endcsname{32.7}
\expandafter\def\csname newEvidence@musique_2hop@gemma4-e4b-w4a16@wt-ci\endcsname{29.1 [25.6, 32.7]}
\expandafter\def\csname newEvidence@musique_2hop@gemma4-e4b-w4a16@null-point\endcsname{29.5}
\expandafter\def\csname newEvidence@musique_2hop@gemma4-e4b-w4a16@null-low\endcsname{26.4}
\expandafter\def\csname newEvidence@musique_2hop@gemma4-e4b-w4a16@null-high\endcsname{32.8}
\expandafter\def\csname newEvidence@musique_2hop@gemma4-e4b-w4a16@null-ci\endcsname{29.5 [26.4, 32.8]}
\expandafter\def\csname newEvidence@musique_2hop@gemma4-e4b-w4a16@lift-point\endcsname{-0.4}
\expandafter\def\csname newEvidence@musique_2hop@gemma4-e4b-w4a16@lift-low\endcsname{-2.0}
\expandafter\def\csname newEvidence@musique_2hop@gemma4-e4b-w4a16@lift-high\endcsname{1.1}
\expandafter\def\csname newEvidence@musique_2hop@gemma4-e4b-w4a16@lift-ci\endcsname{-0.4 [-2.0, 1.1]}
\expandafter\def\csname newEvidence@musique_2hop@gemma4-e4b-w4a16@n_S\endcsname{1000}
\expandafter\def\csname newEvidence@musique_2hop@gemma4-e4b-w4a16@n_wt\endcsname{1000}
\expandafter\def\csname newEvidence@musique_2hop@gemma4-e4b-w4a16@n_binary_eligible_union\endcsname{889}
\expandafter\def\csname newEvidence@musique_2hop@gemma4-e4b-w4a16@flips\endcsname{653}
\expandafter\def\csname newEvidence@musique_2hop@gemma4-e4b-w4a16@helpful_flips\endcsname{190}
\expandafter\def\csname newEvidence@musique_2hop@gemma4-31b-w4a16@S-point\endcsname{1.3}
\expandafter\def\csname newEvidence@musique_2hop@gemma4-31b-w4a16@S-low\endcsname{0.6}
\expandafter\def\csname newEvidence@musique_2hop@gemma4-31b-w4a16@S-high\endcsname{2.0}
\expandafter\def\csname newEvidence@musique_2hop@gemma4-31b-w4a16@S-ci\endcsname{1.3 [0.6, 2.0]}
\expandafter\def\csname newEvidence@musique_2hop@gemma4-31b-w4a16@wt-point\endcsname{33.3}
\expandafter\def\csname newEvidence@musique_2hop@gemma4-31b-w4a16@wt-low\endcsname{6.7}
\expandafter\def\csname newEvidence@musique_2hop@gemma4-31b-w4a16@wt-high\endcsname{63.6}
\expandafter\def\csname newEvidence@musique_2hop@gemma4-31b-w4a16@wt-ci\endcsname{33.3 [6.7, 63.6]}
\expandafter\def\csname newEvidence@musique_2hop@gemma4-31b-w4a16@null-point\endcsname{39.5}
\expandafter\def\csname newEvidence@musique_2hop@gemma4-31b-w4a16@null-low\endcsname{17.2}
\expandafter\def\csname newEvidence@musique_2hop@gemma4-31b-w4a16@null-high\endcsname{65.8}
\expandafter\def\csname newEvidence@musique_2hop@gemma4-31b-w4a16@null-ci\endcsname{39.5 [17.2, 65.8]}
\expandafter\def\csname newEvidence@musique_2hop@gemma4-31b-w4a16@lift-point\endcsname{-6.2}
\expandafter\def\csname newEvidence@musique_2hop@gemma4-31b-w4a16@lift-low\endcsname{-16.2}
\expandafter\def\csname newEvidence@musique_2hop@gemma4-31b-w4a16@lift-high\endcsname{9.4}
\expandafter\def\csname newEvidence@musique_2hop@gemma4-31b-w4a16@lift-ci\endcsname{-6.2 [-16.2, 9.4]}
\expandafter\def\csname newEvidence@musique_2hop@gemma4-31b-w4a16@n_S\endcsname{957}
\expandafter\def\csname newEvidence@musique_2hop@gemma4-31b-w4a16@n_wt\endcsname{953}
\expandafter\def\csname newEvidence@musique_2hop@gemma4-31b-w4a16@n_binary_eligible_union\endcsname{940}
\expandafter\def\csname newEvidence@musique_2hop@gemma4-31b-w4a16@flips\endcsname{15}
\expandafter\def\csname newEvidence@musique_2hop@gemma4-31b-w4a16@helpful_flips\endcsname{5}
\expandafter\def\csname newEvidence@musique_2hop@glm-z1-9b-bnb4@S-point\endcsname{3.7}
\expandafter\def\csname newEvidence@musique_2hop@glm-z1-9b-bnb4@S-low\endcsname{2.4}
\expandafter\def\csname newEvidence@musique_2hop@glm-z1-9b-bnb4@S-high\endcsname{5.0}
\expandafter\def\csname newEvidence@musique_2hop@glm-z1-9b-bnb4@S-ci\endcsname{3.7 [2.4, 5.0]}
\expandafter\def\csname newEvidence@musique_2hop@glm-z1-9b-bnb4@wt-point\endcsname{35.9}
\expandafter\def\csname newEvidence@musique_2hop@glm-z1-9b-bnb4@wt-low\endcsname{20.5}
\expandafter\def\csname newEvidence@musique_2hop@glm-z1-9b-bnb4@wt-high\endcsname{51.9}
\expandafter\def\csname newEvidence@musique_2hop@glm-z1-9b-bnb4@wt-ci\endcsname{35.9 [20.5, 51.9]}
\expandafter\def\csname newEvidence@musique_2hop@glm-z1-9b-bnb4@null-point\endcsname{38.1}
\expandafter\def\csname newEvidence@musique_2hop@glm-z1-9b-bnb4@null-low\endcsname{34.5}
\expandafter\def\csname newEvidence@musique_2hop@glm-z1-9b-bnb4@null-high\endcsname{41.9}
\expandafter\def\csname newEvidence@musique_2hop@glm-z1-9b-bnb4@null-ci\endcsname{38.1 [34.5, 41.9]}
\expandafter\def\csname newEvidence@musique_2hop@glm-z1-9b-bnb4@lift-point\endcsname{-2.2}
\expandafter\def\csname newEvidence@musique_2hop@glm-z1-9b-bnb4@lift-low\endcsname{-17.3}
\expandafter\def\csname newEvidence@musique_2hop@glm-z1-9b-bnb4@lift-high\endcsname{13.5}
\expandafter\def\csname newEvidence@musique_2hop@glm-z1-9b-bnb4@lift-ci\endcsname{-2.2 [-17.3, 13.5]}
\expandafter\def\csname newEvidence@musique_2hop@glm-z1-9b-bnb4@n_S\endcsname{874}
\expandafter\def\csname newEvidence@musique_2hop@glm-z1-9b-bnb4@n_wt\endcsname{788}
\expandafter\def\csname newEvidence@musique_2hop@glm-z1-9b-bnb4@n_binary_eligible_union\endcsname{786}
\expandafter\def\csname newEvidence@musique_2hop@glm-z1-9b-bnb4@flips\endcsname{39}
\expandafter\def\csname newEvidence@musique_2hop@glm-z1-9b-bnb4@helpful_flips\endcsname{14}
\expandafter\def\csname newEvidence@musique_2hop@glm-z1-32b-bnb4@S-point\endcsname{5.6}
\expandafter\def\csname newEvidence@musique_2hop@glm-z1-32b-bnb4@S-low\endcsname{4.1}
\expandafter\def\csname newEvidence@musique_2hop@glm-z1-32b-bnb4@S-high\endcsname{7.2}
\expandafter\def\csname newEvidence@musique_2hop@glm-z1-32b-bnb4@S-ci\endcsname{5.6 [4.1, 7.2]}
\expandafter\def\csname newEvidence@musique_2hop@glm-z1-32b-bnb4@wt-point\endcsname{27.3}
\expandafter\def\csname newEvidence@musique_2hop@glm-z1-32b-bnb4@wt-low\endcsname{14.3}
\expandafter\def\csname newEvidence@musique_2hop@glm-z1-32b-bnb4@wt-high\endcsname{41.2}
\expandafter\def\csname newEvidence@musique_2hop@glm-z1-32b-bnb4@wt-ci\endcsname{27.3 [14.3, 41.2]}
\expandafter\def\csname newEvidence@musique_2hop@glm-z1-32b-bnb4@null-point\endcsname{30.5}
\expandafter\def\csname newEvidence@musique_2hop@glm-z1-32b-bnb4@null-low\endcsname{27.0}
\expandafter\def\csname newEvidence@musique_2hop@glm-z1-32b-bnb4@null-high\endcsname{34.0}
\expandafter\def\csname newEvidence@musique_2hop@glm-z1-32b-bnb4@null-ci\endcsname{30.5 [27.0, 34.0]}
\expandafter\def\csname newEvidence@musique_2hop@glm-z1-32b-bnb4@lift-point\endcsname{-3.3}
\expandafter\def\csname newEvidence@musique_2hop@glm-z1-32b-bnb4@lift-low\endcsname{-15.9}
\expandafter\def\csname newEvidence@musique_2hop@glm-z1-32b-bnb4@lift-high\endcsname{10.3}
\expandafter\def\csname newEvidence@musique_2hop@glm-z1-32b-bnb4@lift-ci\endcsname{-3.3 [-15.9, 10.3]}
\expandafter\def\csname newEvidence@musique_2hop@glm-z1-32b-bnb4@n_S\endcsname{800}
\expandafter\def\csname newEvidence@musique_2hop@glm-z1-32b-bnb4@n_wt\endcsname{767}
\expandafter\def\csname newEvidence@musique_2hop@glm-z1-32b-bnb4@n_binary_eligible_union\endcsname{764}
\expandafter\def\csname newEvidence@musique_2hop@glm-z1-32b-bnb4@flips\endcsname{44}
\expandafter\def\csname newEvidence@musique_2hop@glm-z1-32b-bnb4@helpful_flips\endcsname{12}
\expandafter\def\csname newEvidence@strategyqa@qwen3-4b-fp8@S-point\endcsname{22.2}
\expandafter\def\csname newEvidence@strategyqa@qwen3-4b-fp8@S-low\endcsname{19.5}
\expandafter\def\csname newEvidence@strategyqa@qwen3-4b-fp8@S-high\endcsname{25.0}
\expandafter\def\csname newEvidence@strategyqa@qwen3-4b-fp8@S-ci\endcsname{22.2 [19.5, 25.0]}
\expandafter\def\csname newEvidence@strategyqa@qwen3-4b-fp8@wt-point\endcsname{42.3}
\expandafter\def\csname newEvidence@strategyqa@qwen3-4b-fp8@wt-low\endcsname{36.4}
\expandafter\def\csname newEvidence@strategyqa@qwen3-4b-fp8@wt-high\endcsname{48.3}
\expandafter\def\csname newEvidence@strategyqa@qwen3-4b-fp8@wt-ci\endcsname{42.3 [36.4, 48.3]}
\expandafter\def\csname newEvidence@strategyqa@qwen3-4b-fp8@null-point\endcsname{40.9}
\expandafter\def\csname newEvidence@strategyqa@qwen3-4b-fp8@null-low\endcsname{36.5}
\expandafter\def\csname newEvidence@strategyqa@qwen3-4b-fp8@null-high\endcsname{45.3}
\expandafter\def\csname newEvidence@strategyqa@qwen3-4b-fp8@null-ci\endcsname{40.9 [36.5, 45.3]}
\expandafter\def\csname newEvidence@strategyqa@qwen3-4b-fp8@lift-point\endcsname{1.4}
\expandafter\def\csname newEvidence@strategyqa@qwen3-4b-fp8@lift-low\endcsname{-2.3}
\expandafter\def\csname newEvidence@strategyqa@qwen3-4b-fp8@lift-high\endcsname{5.3}
\expandafter\def\csname newEvidence@strategyqa@qwen3-4b-fp8@lift-ci\endcsname{1.4 [-2.3, 5.3]}
\expandafter\def\csname newEvidence@strategyqa@qwen3-4b-fp8@n_S\endcsname{997}
\expandafter\def\csname newEvidence@strategyqa@qwen3-4b-fp8@n_wt\endcsname{989}
\expandafter\def\csname newEvidence@strategyqa@qwen3-4b-fp8@n_binary_eligible_union\endcsname{989}
\expandafter\def\csname newEvidence@strategyqa@qwen3-4b-fp8@flips\endcsname{352}
\expandafter\def\csname newEvidence@strategyqa@qwen3-4b-fp8@helpful_flips\endcsname{149}
\expandafter\def\csname newEvidence@strategyqa@qwen3-8b-fp8@S-point\endcsname{19.0}
\expandafter\def\csname newEvidence@strategyqa@qwen3-8b-fp8@S-low\endcsname{16.5}
\expandafter\def\csname newEvidence@strategyqa@qwen3-8b-fp8@S-high\endcsname{21.6}
\expandafter\def\csname newEvidence@strategyqa@qwen3-8b-fp8@S-ci\endcsname{19.0 [16.5, 21.6]}
\expandafter\def\csname newEvidence@strategyqa@qwen3-8b-fp8@wt-point\endcsname{39.5}
\expandafter\def\csname newEvidence@strategyqa@qwen3-8b-fp8@wt-low\endcsname{32.8}
\expandafter\def\csname newEvidence@strategyqa@qwen3-8b-fp8@wt-high\endcsname{46.4}
\expandafter\def\csname newEvidence@strategyqa@qwen3-8b-fp8@wt-ci\endcsname{39.5 [32.8, 46.4]}
\expandafter\def\csname newEvidence@strategyqa@qwen3-8b-fp8@null-point\endcsname{35.7}
\expandafter\def\csname newEvidence@strategyqa@qwen3-8b-fp8@null-low\endcsname{30.3}
\expandafter\def\csname newEvidence@strategyqa@qwen3-8b-fp8@null-high\endcsname{41.1}
\expandafter\def\csname newEvidence@strategyqa@qwen3-8b-fp8@null-ci\endcsname{35.7 [30.3, 41.1]}
\expandafter\def\csname newEvidence@strategyqa@qwen3-8b-fp8@lift-point\endcsname{3.8}
\expandafter\def\csname newEvidence@strategyqa@qwen3-8b-fp8@lift-low\endcsname{0.3}
\expandafter\def\csname newEvidence@strategyqa@qwen3-8b-fp8@lift-high\endcsname{7.3}
\expandafter\def\csname newEvidence@strategyqa@qwen3-8b-fp8@lift-ci\endcsname{3.8 [0.3, 7.3]}
\expandafter\def\csname newEvidence@strategyqa@qwen3-8b-fp8@n_S\endcsname{961}
\expandafter\def\csname newEvidence@strategyqa@qwen3-8b-fp8@n_wt\endcsname{992}
\expandafter\def\csname newEvidence@strategyqa@qwen3-8b-fp8@n_binary_eligible_union\endcsname{991}
\expandafter\def\csname newEvidence@strategyqa@qwen3-8b-fp8@flips\endcsname{228}
\expandafter\def\csname newEvidence@strategyqa@qwen3-8b-fp8@helpful_flips\endcsname{90}
\expandafter\def\csname newEvidence@strategyqa@qwen3-14b-fp8@S-point\endcsname{33.9}
\expandafter\def\csname newEvidence@strategyqa@qwen3-14b-fp8@S-low\endcsname{31.0}
\expandafter\def\csname newEvidence@strategyqa@qwen3-14b-fp8@S-high\endcsname{36.8}
\expandafter\def\csname newEvidence@strategyqa@qwen3-14b-fp8@S-ci\endcsname{33.9 [31.0, 36.8]}
\expandafter\def\csname newEvidence@strategyqa@qwen3-14b-fp8@wt-point\endcsname{49.2}
\expandafter\def\csname newEvidence@strategyqa@qwen3-14b-fp8@wt-low\endcsname{44.0}
\expandafter\def\csname newEvidence@strategyqa@qwen3-14b-fp8@wt-high\endcsname{54.6}
\expandafter\def\csname newEvidence@strategyqa@qwen3-14b-fp8@wt-ci\endcsname{49.2 [44.0, 54.6]}
\expandafter\def\csname newEvidence@strategyqa@qwen3-14b-fp8@null-point\endcsname{46.3}
\expandafter\def\csname newEvidence@strategyqa@qwen3-14b-fp8@null-low\endcsname{42.3}
\expandafter\def\csname newEvidence@strategyqa@qwen3-14b-fp8@null-high\endcsname{50.5}
\expandafter\def\csname newEvidence@strategyqa@qwen3-14b-fp8@null-ci\endcsname{46.3 [42.3, 50.5]}
\expandafter\def\csname newEvidence@strategyqa@qwen3-14b-fp8@lift-point\endcsname{2.9}
\expandafter\def\csname newEvidence@strategyqa@qwen3-14b-fp8@lift-low\endcsname{-0.3}
\expandafter\def\csname newEvidence@strategyqa@qwen3-14b-fp8@lift-high\endcsname{6.1}
\expandafter\def\csname newEvidence@strategyqa@qwen3-14b-fp8@lift-ci\endcsname{2.9 [-0.3, 6.1]}
\expandafter\def\csname newEvidence@strategyqa@qwen3-14b-fp8@n_S\endcsname{995}
\expandafter\def\csname newEvidence@strategyqa@qwen3-14b-fp8@n_wt\endcsname{998}
\expandafter\def\csname newEvidence@strategyqa@qwen3-14b-fp8@n_binary_eligible_union\endcsname{998}
\expandafter\def\csname newEvidence@strategyqa@qwen3-14b-fp8@flips\endcsname{384}
\expandafter\def\csname newEvidence@strategyqa@qwen3-14b-fp8@helpful_flips\endcsname{189}
\expandafter\def\csname newEvidence@strategyqa@qwen3-32b-fp8@S-point\endcsname{38.3}
\expandafter\def\csname newEvidence@strategyqa@qwen3-32b-fp8@S-low\endcsname{35.3}
\expandafter\def\csname newEvidence@strategyqa@qwen3-32b-fp8@S-high\endcsname{41.4}
\expandafter\def\csname newEvidence@strategyqa@qwen3-32b-fp8@S-ci\endcsname{38.3 [35.3, 41.4]}
\expandafter\def\csname newEvidence@strategyqa@qwen3-32b-fp8@wt-point\endcsname{54.7}
\expandafter\def\csname newEvidence@strategyqa@qwen3-32b-fp8@wt-low\endcsname{49.8}
\expandafter\def\csname newEvidence@strategyqa@qwen3-32b-fp8@wt-high\endcsname{59.8}
\expandafter\def\csname newEvidence@strategyqa@qwen3-32b-fp8@wt-ci\endcsname{54.7 [49.8, 59.8]}
\expandafter\def\csname newEvidence@strategyqa@qwen3-32b-fp8@null-point\endcsname{53.0}
\expandafter\def\csname newEvidence@strategyqa@qwen3-32b-fp8@null-low\endcsname{48.8}
\expandafter\def\csname newEvidence@strategyqa@qwen3-32b-fp8@null-high\endcsname{57.3}
\expandafter\def\csname newEvidence@strategyqa@qwen3-32b-fp8@null-ci\endcsname{53.0 [48.8, 57.3]}
\expandafter\def\csname newEvidence@strategyqa@qwen3-32b-fp8@lift-point\endcsname{1.7}
\expandafter\def\csname newEvidence@strategyqa@qwen3-32b-fp8@lift-low\endcsname{-0.7}
\expandafter\def\csname newEvidence@strategyqa@qwen3-32b-fp8@lift-high\endcsname{4.1}
\expandafter\def\csname newEvidence@strategyqa@qwen3-32b-fp8@lift-ci\endcsname{1.7 [-0.7, 4.1]}
\expandafter\def\csname newEvidence@strategyqa@qwen3-32b-fp8@n_S\endcsname{1000}
\expandafter\def\csname newEvidence@strategyqa@qwen3-32b-fp8@n_wt\endcsname{1000}
\expandafter\def\csname newEvidence@strategyqa@qwen3-32b-fp8@n_binary_eligible_union\endcsname{1000}
\expandafter\def\csname newEvidence@strategyqa@qwen3-32b-fp8@flips\endcsname{475}
\expandafter\def\csname newEvidence@strategyqa@qwen3-32b-fp8@helpful_flips\endcsname{260}
\expandafter\def\csname newEvidence@strategyqa@gemma4-e2b-w4a16@S-point\endcsname{7.3}
\expandafter\def\csname newEvidence@strategyqa@gemma4-e2b-w4a16@S-low\endcsname{5.7}
\expandafter\def\csname newEvidence@strategyqa@gemma4-e2b-w4a16@S-high\endcsname{9.0}
\expandafter\def\csname newEvidence@strategyqa@gemma4-e2b-w4a16@S-ci\endcsname{7.3 [5.7, 9.0]}
\expandafter\def\csname newEvidence@strategyqa@gemma4-e2b-w4a16@wt-point\endcsname{51.2}
\expandafter\def\csname newEvidence@strategyqa@gemma4-e2b-w4a16@wt-low\endcsname{40.0}
\expandafter\def\csname newEvidence@strategyqa@gemma4-e2b-w4a16@wt-high\endcsname{62.5}
\expandafter\def\csname newEvidence@strategyqa@gemma4-e2b-w4a16@wt-ci\endcsname{51.2 [40.0, 62.5]}
\expandafter\def\csname newEvidence@strategyqa@gemma4-e2b-w4a16@null-point\endcsname{45.4}
\expandafter\def\csname newEvidence@strategyqa@gemma4-e2b-w4a16@null-low\endcsname{40.3}
\expandafter\def\csname newEvidence@strategyqa@gemma4-e2b-w4a16@null-high\endcsname{50.4}
\expandafter\def\csname newEvidence@strategyqa@gemma4-e2b-w4a16@null-ci\endcsname{45.4 [40.3, 50.4]}
\expandafter\def\csname newEvidence@strategyqa@gemma4-e2b-w4a16@lift-point\endcsname{5.9}
\expandafter\def\csname newEvidence@strategyqa@gemma4-e2b-w4a16@lift-low\endcsname{-4.3}
\expandafter\def\csname newEvidence@strategyqa@gemma4-e2b-w4a16@lift-high\endcsname{15.8}
\expandafter\def\csname newEvidence@strategyqa@gemma4-e2b-w4a16@lift-ci\endcsname{5.9 [-4.3, 15.8]}
\expandafter\def\csname newEvidence@strategyqa@gemma4-e2b-w4a16@n_S\endcsname{1000}
\expandafter\def\csname newEvidence@strategyqa@gemma4-e2b-w4a16@n_wt\endcsname{999}
\expandafter\def\csname newEvidence@strategyqa@gemma4-e2b-w4a16@n_binary_eligible_union\endcsname{987}
\expandafter\def\csname newEvidence@strategyqa@gemma4-e2b-w4a16@flips\endcsname{80}
\expandafter\def\csname newEvidence@strategyqa@gemma4-e2b-w4a16@helpful_flips\endcsname{41}
\expandafter\def\csname newEvidence@strategyqa@gemma4-e4b-w4a16@S-point\endcsname{25.8}
\expandafter\def\csname newEvidence@strategyqa@gemma4-e4b-w4a16@S-low\endcsname{23.1}
\expandafter\def\csname newEvidence@strategyqa@gemma4-e4b-w4a16@S-high\endcsname{28.5}
\expandafter\def\csname newEvidence@strategyqa@gemma4-e4b-w4a16@S-ci\endcsname{25.8 [23.1, 28.5]}
\expandafter\def\csname newEvidence@strategyqa@gemma4-e4b-w4a16@wt-point\endcsname{30.6}
\expandafter\def\csname newEvidence@strategyqa@gemma4-e4b-w4a16@wt-low\endcsname{24.8}
\expandafter\def\csname newEvidence@strategyqa@gemma4-e4b-w4a16@wt-high\endcsname{36.5}
\expandafter\def\csname newEvidence@strategyqa@gemma4-e4b-w4a16@wt-ci\endcsname{30.6 [24.8, 36.5]}
\expandafter\def\csname newEvidence@strategyqa@gemma4-e4b-w4a16@null-point\endcsname{30.0}
\expandafter\def\csname newEvidence@strategyqa@gemma4-e4b-w4a16@null-low\endcsname{25.2}
\expandafter\def\csname newEvidence@strategyqa@gemma4-e4b-w4a16@null-high\endcsname{34.8}
\expandafter\def\csname newEvidence@strategyqa@gemma4-e4b-w4a16@null-ci\endcsname{30.0 [25.2, 34.8]}
\expandafter\def\csname newEvidence@strategyqa@gemma4-e4b-w4a16@lift-point\endcsname{0.6}
\expandafter\def\csname newEvidence@strategyqa@gemma4-e4b-w4a16@lift-low\endcsname{-2.6}
\expandafter\def\csname newEvidence@strategyqa@gemma4-e4b-w4a16@lift-high\endcsname{3.8}
\expandafter\def\csname newEvidence@strategyqa@gemma4-e4b-w4a16@lift-ci\endcsname{0.6 [-2.6, 3.8]}
\expandafter\def\csname newEvidence@strategyqa@gemma4-e4b-w4a16@n_S\endcsname{1000}
\expandafter\def\csname newEvidence@strategyqa@gemma4-e4b-w4a16@n_wt\endcsname{1000}
\expandafter\def\csname newEvidence@strategyqa@gemma4-e4b-w4a16@n_binary_eligible_union\endcsname{972}
\expandafter\def\csname newEvidence@strategyqa@gemma4-e4b-w4a16@flips\endcsname{255}
\expandafter\def\csname newEvidence@strategyqa@gemma4-e4b-w4a16@helpful_flips\endcsname{78}
\expandafter\def\csname newEvidence@strategyqa@gemma4-31b-w4a16@S-point\endcsname{17.3}
\expandafter\def\csname newEvidence@strategyqa@gemma4-31b-w4a16@S-low\endcsname{14.9}
\expandafter\def\csname newEvidence@strategyqa@gemma4-31b-w4a16@S-high\endcsname{19.8}
\expandafter\def\csname newEvidence@strategyqa@gemma4-31b-w4a16@S-ci\endcsname{17.3 [14.9, 19.8]}
\expandafter\def\csname newEvidence@strategyqa@gemma4-31b-w4a16@wt-point\endcsname{44.1}
\expandafter\def\csname newEvidence@strategyqa@gemma4-31b-w4a16@wt-low\endcsname{36.7}
\expandafter\def\csname newEvidence@strategyqa@gemma4-31b-w4a16@wt-high\endcsname{51.3}
\expandafter\def\csname newEvidence@strategyqa@gemma4-31b-w4a16@wt-ci\endcsname{44.1 [36.7, 51.3]}
\expandafter\def\csname newEvidence@strategyqa@gemma4-31b-w4a16@null-point\endcsname{33.0}
\expandafter\def\csname newEvidence@strategyqa@gemma4-31b-w4a16@null-low\endcsname{28.1}
\expandafter\def\csname newEvidence@strategyqa@gemma4-31b-w4a16@null-high\endcsname{38.0}
\expandafter\def\csname newEvidence@strategyqa@gemma4-31b-w4a16@null-ci\endcsname{33.0 [28.1, 38.0]}
\expandafter\def\csname newEvidence@strategyqa@gemma4-31b-w4a16@lift-point\endcsname{11.1}
\expandafter\def\csname newEvidence@strategyqa@gemma4-31b-w4a16@lift-low\endcsname{5.7}
\expandafter\def\csname newEvidence@strategyqa@gemma4-31b-w4a16@lift-high\endcsname{16.5}
\expandafter\def\csname newEvidence@strategyqa@gemma4-31b-w4a16@lift-ci\endcsname{11.1 [5.7, 16.5]}
\expandafter\def\csname newEvidence@strategyqa@gemma4-31b-w4a16@n_S\endcsname{998}
\expandafter\def\csname newEvidence@strategyqa@gemma4-31b-w4a16@n_wt\endcsname{985}
\expandafter\def\csname newEvidence@strategyqa@gemma4-31b-w4a16@n_binary_eligible_union\endcsname{984}
\expandafter\def\csname newEvidence@strategyqa@gemma4-31b-w4a16@flips\endcsname{202}
\expandafter\def\csname newEvidence@strategyqa@gemma4-31b-w4a16@helpful_flips\endcsname{89}
\expandafter\def\csname newEvidence@strategyqa@glm-z1-9b-bnb4@S-point\endcsname{10.1}
\expandafter\def\csname newEvidence@strategyqa@glm-z1-9b-bnb4@S-low\endcsname{8.2}
\expandafter\def\csname newEvidence@strategyqa@glm-z1-9b-bnb4@S-high\endcsname{12.1}
\expandafter\def\csname newEvidence@strategyqa@glm-z1-9b-bnb4@S-ci\endcsname{10.1 [8.2, 12.1]}
\expandafter\def\csname newEvidence@strategyqa@glm-z1-9b-bnb4@wt-point\endcsname{42.9}
\expandafter\def\csname newEvidence@strategyqa@glm-z1-9b-bnb4@wt-low\endcsname{33.3}
\expandafter\def\csname newEvidence@strategyqa@glm-z1-9b-bnb4@wt-high\endcsname{52.5}
\expandafter\def\csname newEvidence@strategyqa@glm-z1-9b-bnb4@wt-ci\endcsname{42.9 [33.3, 52.5]}
\expandafter\def\csname newEvidence@strategyqa@glm-z1-9b-bnb4@null-point\endcsname{39.9}
\expandafter\def\csname newEvidence@strategyqa@glm-z1-9b-bnb4@null-low\endcsname{33.7}
\expandafter\def\csname newEvidence@strategyqa@glm-z1-9b-bnb4@null-high\endcsname{45.9}
\expandafter\def\csname newEvidence@strategyqa@glm-z1-9b-bnb4@null-ci\endcsname{39.9 [33.7, 45.9]}
\expandafter\def\csname newEvidence@strategyqa@glm-z1-9b-bnb4@lift-point\endcsname{3.0}
\expandafter\def\csname newEvidence@strategyqa@glm-z1-9b-bnb4@lift-low\endcsname{-5.0}
\expandafter\def\csname newEvidence@strategyqa@glm-z1-9b-bnb4@lift-high\endcsname{10.7}
\expandafter\def\csname newEvidence@strategyqa@glm-z1-9b-bnb4@lift-ci\endcsname{3.0 [-5.0, 10.7]}
\expandafter\def\csname newEvidence@strategyqa@glm-z1-9b-bnb4@n_S\endcsname{982}
\expandafter\def\csname newEvidence@strategyqa@glm-z1-9b-bnb4@n_wt\endcsname{980}
\expandafter\def\csname newEvidence@strategyqa@glm-z1-9b-bnb4@n_binary_eligible_union\endcsname{980}
\expandafter\def\csname newEvidence@strategyqa@glm-z1-9b-bnb4@flips\endcsname{112}
\expandafter\def\csname newEvidence@strategyqa@glm-z1-9b-bnb4@helpful_flips\endcsname{48}
\expandafter\def\csname newEvidence@strategyqa@glm-z1-32b-bnb4@S-point\endcsname{7.2}
\expandafter\def\csname newEvidence@strategyqa@glm-z1-32b-bnb4@S-low\endcsname{5.5}
\expandafter\def\csname newEvidence@strategyqa@glm-z1-32b-bnb4@S-high\endcsname{8.9}
\expandafter\def\csname newEvidence@strategyqa@glm-z1-32b-bnb4@S-ci\endcsname{7.2 [5.5, 8.9]}
\expandafter\def\csname newEvidence@strategyqa@glm-z1-32b-bnb4@wt-point\endcsname{45.5}
\expandafter\def\csname newEvidence@strategyqa@glm-z1-32b-bnb4@wt-low\endcsname{34.7}
\expandafter\def\csname newEvidence@strategyqa@glm-z1-32b-bnb4@wt-high\endcsname{56.2}
\expandafter\def\csname newEvidence@strategyqa@glm-z1-32b-bnb4@wt-ci\endcsname{45.5 [34.7, 56.2]}
\expandafter\def\csname newEvidence@strategyqa@glm-z1-32b-bnb4@null-point\endcsname{43.5}
\expandafter\def\csname newEvidence@strategyqa@glm-z1-32b-bnb4@null-low\endcsname{35.6}
\expandafter\def\csname newEvidence@strategyqa@glm-z1-32b-bnb4@null-high\endcsname{51.2}
\expandafter\def\csname newEvidence@strategyqa@glm-z1-32b-bnb4@null-ci\endcsname{43.5 [35.6, 51.2]}
\expandafter\def\csname newEvidence@strategyqa@glm-z1-32b-bnb4@lift-point\endcsname{2.0}
\expandafter\def\csname newEvidence@strategyqa@glm-z1-32b-bnb4@lift-low\endcsname{-5.6}
\expandafter\def\csname newEvidence@strategyqa@glm-z1-32b-bnb4@lift-high\endcsname{9.4}
\expandafter\def\csname newEvidence@strategyqa@glm-z1-32b-bnb4@lift-ci\endcsname{2.0 [-5.6, 9.4]}
\expandafter\def\csname newEvidence@strategyqa@glm-z1-32b-bnb4@n_S\endcsname{987}
\expandafter\def\csname newEvidence@strategyqa@glm-z1-32b-bnb4@n_wt\endcsname{990}
\expandafter\def\csname newEvidence@strategyqa@glm-z1-32b-bnb4@n_binary_eligible_union\endcsname{990}
\expandafter\def\csname newEvidence@strategyqa@glm-z1-32b-bnb4@flips\endcsname{88}
\expandafter\def\csname newEvidence@strategyqa@glm-z1-32b-bnb4@helpful_flips\endcsname{40}
\expandafter\def\csname newEvidence@musique_2hop@deepseek-v4-pro@S-point\endcsname{66.8}
\expandafter\def\csname newEvidence@musique_2hop@deepseek-v4-pro@S-low\endcsname{61.2}
\expandafter\def\csname newEvidence@musique_2hop@deepseek-v4-pro@S-high\endcsname{72.4}
\expandafter\def\csname newEvidence@musique_2hop@deepseek-v4-pro@S-ci\endcsname{66.8 [61.2, 72.4]}
\expandafter\def\csname newEvidence@musique_2hop@deepseek-v4-pro@wt-point\endcsname{28.7}
\expandafter\def\csname newEvidence@musique_2hop@deepseek-v4-pro@wt-low\endcsname{21.9}
\expandafter\def\csname newEvidence@musique_2hop@deepseek-v4-pro@wt-high\endcsname{35.8}
\expandafter\def\csname newEvidence@musique_2hop@deepseek-v4-pro@wt-ci\endcsname{28.7 [21.9, 35.8]}
\expandafter\def\csname newEvidence@musique_2hop@deepseek-v4-pro@null-point\endcsname{24.1}
\expandafter\def\csname newEvidence@musique_2hop@deepseek-v4-pro@null-low\endcsname{18.8}
\expandafter\def\csname newEvidence@musique_2hop@deepseek-v4-pro@null-high\endcsname{29.6}
\expandafter\def\csname newEvidence@musique_2hop@deepseek-v4-pro@null-ci\endcsname{24.1 [18.8, 29.6]}
\expandafter\def\csname newEvidence@musique_2hop@deepseek-v4-pro@lift-point\endcsname{4.6}
\expandafter\def\csname newEvidence@musique_2hop@deepseek-v4-pro@lift-low\endcsname{1.5}
\expandafter\def\csname newEvidence@musique_2hop@deepseek-v4-pro@lift-high\endcsname{7.8}
\expandafter\def\csname newEvidence@musique_2hop@deepseek-v4-pro@lift-ci\endcsname{4.6 [1.5, 7.8]}
\expandafter\def\csname newEvidence@musique_2hop@deepseek-v4-pro@n_S\endcsname{250}
\expandafter\def\csname newEvidence@musique_2hop@deepseek-v4-pro@n_wt\endcsname{250}
\expandafter\def\csname newEvidence@musique_2hop@deepseek-v4-pro@n_binary_eligible_union\endcsname{250}
\expandafter\def\csname newEvidence@musique_2hop@deepseek-v4-pro@flips\endcsname{167}
\expandafter\def\csname newEvidence@musique_2hop@deepseek-v4-pro@helpful_flips\endcsname{48}
\expandafter\def\csname newEvidence@strategyqa@deepseek-v4-pro@S-point\endcsname{70.0}
\expandafter\def\csname newEvidence@strategyqa@deepseek-v4-pro@S-low\endcsname{64.4}
\expandafter\def\csname newEvidence@strategyqa@deepseek-v4-pro@S-high\endcsname{75.6}
\expandafter\def\csname newEvidence@strategyqa@deepseek-v4-pro@S-ci\endcsname{70.0 [64.4, 75.6]}
\expandafter\def\csname newEvidence@strategyqa@deepseek-v4-pro@wt-point\endcsname{47.7}
\expandafter\def\csname newEvidence@strategyqa@deepseek-v4-pro@wt-low\endcsname{40.4}
\expandafter\def\csname newEvidence@strategyqa@deepseek-v4-pro@wt-high\endcsname{55.2}
\expandafter\def\csname newEvidence@strategyqa@deepseek-v4-pro@wt-ci\endcsname{47.7 [40.4, 55.2]}
\expandafter\def\csname newEvidence@strategyqa@deepseek-v4-pro@null-point\endcsname{48.9}
\expandafter\def\csname newEvidence@strategyqa@deepseek-v4-pro@null-low\endcsname{42.1}
\expandafter\def\csname newEvidence@strategyqa@deepseek-v4-pro@null-high\endcsname{55.7}
\expandafter\def\csname newEvidence@strategyqa@deepseek-v4-pro@null-ci\endcsname{48.9 [42.1, 55.7]}
\expandafter\def\csname newEvidence@strategyqa@deepseek-v4-pro@lift-point\endcsname{-1.1}
\expandafter\def\csname newEvidence@strategyqa@deepseek-v4-pro@lift-low\endcsname{-3.7}
\expandafter\def\csname newEvidence@strategyqa@deepseek-v4-pro@lift-high\endcsname{1.5}
\expandafter\def\csname newEvidence@strategyqa@deepseek-v4-pro@lift-ci\endcsname{-1.1 [-3.7, 1.5]}
\expandafter\def\csname newEvidence@strategyqa@deepseek-v4-pro@n_S\endcsname{250}
\expandafter\def\csname newEvidence@strategyqa@deepseek-v4-pro@n_wt\endcsname{249}
\expandafter\def\csname newEvidence@strategyqa@deepseek-v4-pro@n_binary_eligible_union\endcsname{249}
\expandafter\def\csname newEvidence@strategyqa@deepseek-v4-pro@flips\endcsname{176}
\expandafter\def\csname newEvidence@strategyqa@deepseek-v4-pro@helpful_flips\endcsname{84}
\expandafter\def\csname newEvidence@musique_2hop@minimax-m3@S-point\endcsname{66.4}
\expandafter\def\csname newEvidence@musique_2hop@minimax-m3@S-low\endcsname{60.2}
\expandafter\def\csname newEvidence@musique_2hop@minimax-m3@S-high\endcsname{72.1}
\expandafter\def\csname newEvidence@musique_2hop@minimax-m3@S-ci\endcsname{66.4 [60.2, 72.1]}
\expandafter\def\csname newEvidence@musique_2hop@minimax-m3@wt-point\endcsname{27.1}
\expandafter\def\csname newEvidence@musique_2hop@minimax-m3@wt-low\endcsname{20.3}
\expandafter\def\csname newEvidence@musique_2hop@minimax-m3@wt-high\endcsname{34.1}
\expandafter\def\csname newEvidence@musique_2hop@minimax-m3@wt-ci\endcsname{27.1 [20.3, 34.1]}
\expandafter\def\csname newEvidence@musique_2hop@minimax-m3@null-point\endcsname{25.8}
\expandafter\def\csname newEvidence@musique_2hop@minimax-m3@null-low\endcsname{19.9}
\expandafter\def\csname newEvidence@musique_2hop@minimax-m3@null-high\endcsname{32.0}
\expandafter\def\csname newEvidence@musique_2hop@minimax-m3@null-ci\endcsname{25.8 [19.9, 32.0]}
\expandafter\def\csname newEvidence@musique_2hop@minimax-m3@lift-point\endcsname{1.2}
\expandafter\def\csname newEvidence@musique_2hop@minimax-m3@lift-low\endcsname{-2.0}
\expandafter\def\csname newEvidence@musique_2hop@minimax-m3@lift-high\endcsname{4.5}
\expandafter\def\csname newEvidence@musique_2hop@minimax-m3@lift-ci\endcsname{1.2 [-2.0, 4.5]}
\expandafter\def\csname newEvidence@musique_2hop@minimax-m3@n_S\endcsname{244}
\expandafter\def\csname newEvidence@musique_2hop@minimax-m3@n_wt\endcsname{238}
\expandafter\def\csname newEvidence@musique_2hop@minimax-m3@n_binary_eligible_union\endcsname{238}
\expandafter\def\csname newEvidence@musique_2hop@minimax-m3@flips\endcsname{170}
\expandafter\def\csname newEvidence@musique_2hop@minimax-m3@helpful_flips\endcsname{46}
\expandafter\def\csname newEvidence@strategyqa@minimax-m3@S-point\endcsname{53.0}
\expandafter\def\csname newEvidence@strategyqa@minimax-m3@S-low\endcsname{46.6}
\expandafter\def\csname newEvidence@strategyqa@minimax-m3@S-high\endcsname{59.0}
\expandafter\def\csname newEvidence@strategyqa@minimax-m3@S-ci\endcsname{53.0 [46.6, 59.0]}
\expandafter\def\csname newEvidence@strategyqa@minimax-m3@wt-point\endcsname{30.4}
\expandafter\def\csname newEvidence@strategyqa@minimax-m3@wt-low\endcsname{22.6}
\expandafter\def\csname newEvidence@strategyqa@minimax-m3@wt-high\endcsname{38.5}
\expandafter\def\csname newEvidence@strategyqa@minimax-m3@wt-ci\endcsname{30.4 [22.6, 38.5]}
\expandafter\def\csname newEvidence@strategyqa@minimax-m3@null-point\endcsname{30.4}
\expandafter\def\csname newEvidence@strategyqa@minimax-m3@null-low\endcsname{23.8}
\expandafter\def\csname newEvidence@strategyqa@minimax-m3@null-high\endcsname{37.3}
\expandafter\def\csname newEvidence@strategyqa@minimax-m3@null-ci\endcsname{30.4 [23.8, 37.3]}
\expandafter\def\csname newEvidence@strategyqa@minimax-m3@lift-point\endcsname{-0.1}
\expandafter\def\csname newEvidence@strategyqa@minimax-m3@lift-low\endcsname{-4.6}
\expandafter\def\csname newEvidence@strategyqa@minimax-m3@lift-high\endcsname{4.3}
\expandafter\def\csname newEvidence@strategyqa@minimax-m3@lift-ci\endcsname{-0.1 [-4.6, 4.3]}
\expandafter\def\csname newEvidence@strategyqa@minimax-m3@n_S\endcsname{249}
\expandafter\def\csname newEvidence@strategyqa@minimax-m3@n_wt\endcsname{250}
\expandafter\def\csname newEvidence@strategyqa@minimax-m3@n_binary_eligible_union\endcsname{249}
\expandafter\def\csname newEvidence@strategyqa@minimax-m3@flips\endcsname{135}
\expandafter\def\csname newEvidence@strategyqa@minimax-m3@helpful_flips\endcsname{41}
 \expandafter\def\csname newEvidence@musique_2hop@open-weight-median@S-point\endcsname{22.4}
\expandafter\def\csname newEvidence@musique_2hop@open-weight-median@wt-point\endcsname{36.9}
\expandafter\def\csname newEvidence@musique_2hop@open-weight-median@lift-point\endcsname{-0.4}
\expandafter\def\csname newEvidence@strategyqa@open-weight-median@S-point\endcsname{19.0}
\expandafter\def\csname newEvidence@strategyqa@open-weight-median@wt-point\endcsname{44.1}
\expandafter\def\csname newEvidence@strategyqa@open-weight-median@lift-point\endcsname{2.9}
 \providecommand{\unaidedAcc}[2]{\csname unaidedAcc@#1@#2\endcsname}
\expandafter\def\csname unaidedAcc@musique_2hop@qwen3-4b-fp8\endcsname{58.6}
\expandafter\def\csname unaidedAcc@musique_2hop@qwen3-8b-fp8\endcsname{63.6}
\expandafter\def\csname unaidedAcc@musique_2hop@qwen3-14b-fp8\endcsname{63.1}
\expandafter\def\csname unaidedAcc@musique_2hop@qwen3-32b-fp8\endcsname{66.3}
\expandafter\def\csname unaidedAcc@musique_2hop@gemma4-e2b-w4a16\endcsname{51.2}
\expandafter\def\csname unaidedAcc@musique_2hop@gemma4-e4b-w4a16\endcsname{61.6}
\expandafter\def\csname unaidedAcc@musique_2hop@gemma4-31b-w4a16\endcsname{82.8}
\expandafter\def\csname unaidedAcc@musique_2hop@glm-z1-9b-bnb4\endcsname{58.1}
\expandafter\def\csname unaidedAcc@musique_2hop@glm-z1-32b-bnb4\endcsname{65.9}
\expandafter\def\csname unaidedAcc@strategyqa@qwen3-4b-fp8\endcsname{71.7}
\expandafter\def\csname unaidedAcc@strategyqa@qwen3-8b-fp8\endcsname{76.4}
\expandafter\def\csname unaidedAcc@strategyqa@qwen3-14b-fp8\endcsname{75.8}
\expandafter\def\csname unaidedAcc@strategyqa@qwen3-32b-fp8\endcsname{81.6}
\expandafter\def\csname unaidedAcc@strategyqa@gemma4-e2b-w4a16\endcsname{61.7}
\expandafter\def\csname unaidedAcc@strategyqa@gemma4-e4b-w4a16\endcsname{74.3}
\expandafter\def\csname unaidedAcc@strategyqa@gemma4-31b-w4a16\endcsname{83.0}
\expandafter\def\csname unaidedAcc@strategyqa@glm-z1-9b-bnb4\endcsname{73.8}
\expandafter\def\csname unaidedAcc@strategyqa@glm-z1-32b-bnb4\endcsname{79.7}
\expandafter\def\csname unaidedAcc@musique_2hop@deepseek-v4-pro\endcsname{78.4}
\expandafter\def\csname unaidedAcc@strategyqa@deepseek-v4-pro\endcsname{84.7}
\expandafter\def\csname unaidedAcc@musique_2hop@minimax-m3\endcsname{81.7}
\expandafter\def\csname unaidedAcc@strategyqa@minimax-m3\endcsname{82.4}
\expandafter\def\csname unaidedAcc@musique_2hop@open-weight-median\endcsname{63.1}
\expandafter\def\csname unaidedAcc@musique_2hop@open-weight-min\endcsname{51.2}
\expandafter\def\csname unaidedAcc@musique_2hop@open-weight-max\endcsname{82.8}
\expandafter\def\csname unaidedAcc@strategyqa@open-weight-median\endcsname{75.8}
\expandafter\def\csname unaidedAcc@strategyqa@open-weight-min\endcsname{61.7}
\expandafter\def\csname unaidedAcc@strategyqa@open-weight-max\endcsname{83.0}
 
\begin{document}

\maketitle

\begin{abstract}
Reasoning language models that can call tools must decide during inference whether to answer unaided or delegate. Any proposal for such self-reflection mechanism has to answer three questions: \emph{where does the reflective signal come from} (verbal reports, output distributions, hidden states, a separate predictor), \emph{how it is presented to the model} (numerical prediction, confidence token, prompt injection), and \emph{whether it changes the model's subsequent action}. We isolate and study the third question. We intervene on otherwise identical reasoning trajectories by inserting, at a fixed point after the same prompt and reasoning prefix, a single first-person sentence expressing either confidence or doubt. The model then continues reasoning and chooses whether to answer directly or call a tool. Comparing these counterfactual continuations measures the causal effect of the reflective signal on delegation while holding the preceding reasoning trajectory fixed. We define this change in behavioral response as Nudgeability. It is a property of large language models, measured along two dimensions: \emph{sensitivity}, how strongly confidence and doubt change delegation rates, and \emph{targeting}, whether delegation increases for problems the model cannot solve unaided and decreases for those it can.

Across nine small-to-medium open-weight reasoning models from three families (Qwen, Gemma, and GLM) and two tasks, models are consistently sensitive to the intervention: doubt increases delegation and confidence decreases it, with a median confidence-to-doubt swing of 20.6 percentage points. The larger provider-served models exhibit swings of 53 to 70 percentage points. This responsiveness, however, is poorly targeted. Although a median 42\% of induced behavioral flips are well-targeted, this represents only a +2 percentage-point median lift over a random-selection baseline. Thus confidence language provides a strong control surface for delegation, but current models seem to use such signals only weakly in accordance with their actual unaided competence. Nudgeability gives us a simple post-training-free way to evaluate both, sensitivity and targeting, as endogenous self-reflection mechanisms mature.
\end{abstract}

\section{Introduction}
\label{sec:introduction}

Reasoning language models increasingly operate in settings where they can either answer from their own computation or call a tool for help. Tool use can extend a model's effective capabilities, but the model must first decide when that help is needed. Answering unaided can fail when the model cannot solve the problem, while delegating a problem it can already solve is unnecessary. In systems that interleave reasoning and action, this is not merely a routing decision made before inference: it can arise inside an ongoing reasoning trajectory, after the model has formed a tentative answer but before it commits to an action \citep{yao2023react,wu2026call}. A useful reflective mechanism must therefore change behavior on the right problems, not merely increase or decrease tool use on average.

Proposals for self-reflection in language models combine several distinct questions. First, \emph{where does the reflective signal come from?} Systems such as Reflexion and Self-Refine use model-generated verbal feedback \citep{shinn2023reflexion,madaan2023selfrefine}; other approaches estimate uncertainty from output distributions, hidden states, or a separate predictor \citep{kadavath2022know,farquhar2024semantic,kossen2024semantic,ghasemabadi2025gnosis}. Second, \emph{how is that signal presented to the model?} The same underlying estimate could be supplied as ordinary language, a learned confidence token, or a continuous representation returned to the reasoning process \citep{chuang2025confidence,guo2026uncertainty,zhang2025softthinking,li2025metacognitive}. Third, \emph{does receiving the signal change what the model subsequently does?} These questions concern, respectively, the source, representation, and behavioral effect of reflection. They need not have the same answer: an informative signal may be behaviorally inert, whereas an ungrounded statement may strongly steer a model.

This separation distinguishes our use of confidence language from work that trains models to express uncertainty explicitly \citep{guo2026uncertainty}, where the confidence expression is itself evaluated against answer accuracy. We instead supply the signal experimentally and measure whether it changes a later delegation decision; accuracy labels enter only afterward, to evaluate whether the induced action changes are targeted to problems on which help is needed.
We take the delegation decision itself, rather than downstream accuracy, as the outcome. The information and computation available to a model unaided are necessarily bounded, so delegating under uncertainty is a more robust route to reliability than improving unaided answers alone; \citet{wu2026call} likewise evaluate the call/no-call decision directly.

\begin{figure*}[t]
  \centering
  \includegraphics[width=\textwidth]{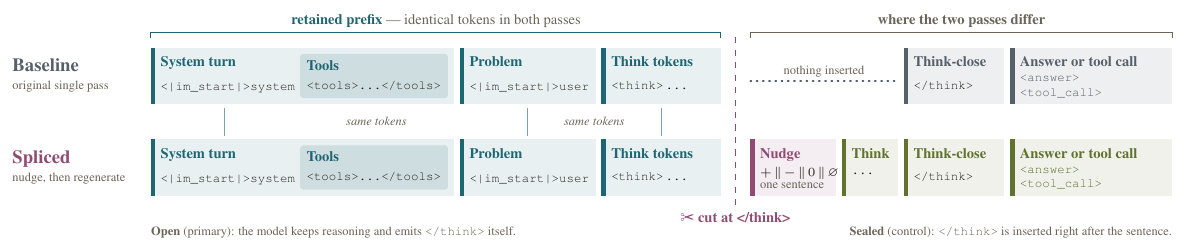}
  \caption{Anatomy of the intervention prompt. The system prompt, tool description, problem, and generated think tokens form the prefix preceding the intervention. The confidence or doubt nudge is spliced immediately before the original think-close boundary, after which the continuation is regenerated. In the primary open-continuation condition, the model may continue reasoning before committing to an answer or tool call. The upper row is the original uninterrupted pass the prefix was taken from.}
  \label{fig:prompt-anatomy}
\end{figure*}

The supplied intervention is a single first-person sentence expressing either confidence or doubt. The insertion occurs at the same boundary after an identical prompt and reasoning prefix, and the model is then allowed to continue reasoning before choosing whether to answer directly or call a tool. Figure~\ref{fig:prompt-anatomy} shows the prompt components and splice boundary. Because the two arms share everything preceding the intervention, their difference in delegation isolates the effect of the inserted sentence rather than differences in the problem, prompt, or reasoning that led up to it.

We use an ordinary English sentence because tokenized natural language is the common interface shared by all models in our evaluation, and by language models more generally. The signal is represented in the same input space in which these models already continue reasoning, so the intervention requires no learned confidence token, architecture-specific activation access, fine-tuning, or other post-training modification. This choice favors portability and experimental control; we do not claim that prose is the best representation for an eventual endogenous signal.

We call this behavioral property \emph{Nudgeability}. Nudgeability has two dimensions. \emph{Sensitivity} measures how strongly the intervention changes delegation: does doubt move the model toward tool use, and confidence toward answering unaided? \emph{Targeting} measures whether those changes agree with the model's demonstrated unaided competence: does delegation increase on problems the model fails and decrease on problems it solves? An intervention can produce a large aggregate shift by moving many decisions indiscriminately, without improving the selection of problems on which help is sought. We therefore assess targeting both by the fraction of induced action changes that are competence-appropriate and by its lift over a random-selection baseline. Targeting is thus related to abstention, where a model should refrain from answering when it is likely to be wrong \citep{madhusudhan2025abstention,wen2025abstention}; delegation instead continues the attempt with outside help.

Our intervention treats the reasoning trace as an experimental interface. Prior work shows that inserting, revising, or perturbing reasoning tokens can change model outputs, while also cautioning that displayed reasoning need not faithfully report the factors that caused an answer \citep{wu2025thinking,lanham2023faithfulness,turpin2023unfaithful}. Model providers also expose, summarize, encrypt, or validate reasoning channels differently from ordinary conversational text, treating reasoning blocks as a privileged\footnote{We mean privileged \emph{influence} on a later action, not privileged \emph{access} to self-information (Section~\ref{sec:implications}). Providers may also restrict these channels for other reasons, such as preventing edited reasoning from bypassing safeguards or limiting distillation, whose relative importance is not publicly documented; we test only the causal-influence claim.} part of the model's output \citep{openai2025modelspec,google2026thinking,anthropic2026thinking}. They motivate studying the trace as a distinct intervention site. Unlike these works, we vary a single semantic feature at a fixed boundary of an observed prefix, measure a discrete delegation decision on paired counterfactual arms, and score the induced changes against unaided competence; such simple controlled interventions can reveal stable behavioral structure without a complete mechanistic theory \citep{simon2026theory}.

The experiment makes no claim that the inserted sentence is a truthful self-assessment or that the model has privileged access to its own uncertainty. The confidence signal is exogenous, and its first-person form is held fixed. Consequently, Nudgeability does not measure the quality of a reflective signal's source, nor does it establish introspection in the stronger sense of privileged self-knowledge \citep{song2025privileged}. Instead, it measures whether a signal, once placed inside an ongoing reasoning trajectory, acts as a control surface for delegation. This separation is useful for the development of endogenous systems: before asking whether a model can produce a reliable reflective signal, we can test whether a candidate representation would influence the intended downstream decision at all.

Our contributions are threefold:
\begin{enumerate}[leftmargin=*,itemsep=2pt,topsep=3pt]
  \item We separate reflective mechanisms into source, representation, and behavioral-effect questions, and introduce Nudgeability as a property of how a model's action responds to a supplied reflective signal.
  \item We give a paired, post-training-free framework for measuring Nudgeability while holding the preceding reasoning trajectory fixed. The framework distinguishes sensitivity from targeting and evaluates targeting against a matched random-selection baseline.
  \item We evaluate nine open-weight reasoning models from the Qwen, Gemma, and GLM families on two tasks, together with larger provider-hosted DeepSeek and MiniMax models. Confidence language strongly changes delegation, with a median confidence-to-doubt swing of 20.6 percentage points and swings of 53--70 points for the provider-hosted models. The induced changes are only weakly targeted to unaided competence: a median 42\% of flips are well-targeted, only +2 percentage points above the random-selection baseline.
\end{enumerate}

\section{Nudgeability Framework}
\label{sec:framework}

In this section, we formalize Nudgeability for a model, task, and tool configuration and define the two key metrics: \emph{sensitivity} and \emph{targeting}. Later, we briefly touch on the control experiments we performed. We have performed a more exhaustive comparison of our framework with related lines of research in Appendix~\ref{app:related-work}.

\paragraph{Setting and action space.}
\label{sec:framework-setting}

Let $M$ be a reasoning language model and let $i$ denote a problem with a verifiable answer. A tool $T$ may be described in the prompt and made available through the model's native tool-calling interface. Let $c\in\mathcal{C}$ index the run condition that produces an action. In the primary experiment, $c=\mathrm{NT}$ denotes the uninterrupted no-tool base, $c=B$ the uninterrupted with-tool base, and $c\in\{\none,+,-,0\}$ the sentence-less, confidence, doubt, and neutral splice conditions, respectively. Supporting controls extend $\mathcal{C}$ with condition-specific labels. After a reasoning segment, the model produces a committed action
\begin{equation}
  A_i^c \in \{\self,\tool,\other\},
  \label{eq:action-space}
\end{equation}
The action is read from the committed continuation after the model-specific think boundary. The tool call itself is never executed, in any condition; we record only the committed action. The category \other{} records a completed but malformed commitment. A generation that reaches its token budget before committing is treated as truncated and removed from the observation set.

Two uninterrupted runs, the \emph{no-tool base} and the \emph{with-tool base}, provide the reference behavior. The no-tool base measures unaided competence. We set
\begin{equation}
  s_i =
  \begin{cases}
    1, & \text{if the no-tool answer to $i$ is correct},\\
    0, & \text{if the no-tool answer to $i$ is incorrect}.
  \end{cases}
  \label{eq:competence-label}
\end{equation}
Problems with an unknown or truncated no-tool answer have no competence label and are excluded from estimators that require $s_i$. This label is an operational, single-run measure of unaided performance; it is not a claim about a model's latent ability under arbitrary resampling. The with-tool base yields the uninterrupted action $A_i^B$, the reference against which intervention-induced action changes are measured.

Each estimator uses its own pairwise-complete retained set. If an estimator requires several conditions and any required generation is truncated or otherwise unreadable, problem $i$ is omitted from that estimator. Consequently, every reported sample size is a retained count for a particular contrast rather than a universal dataset size.

\paragraph{Paired reasoning intervention.}
\label{sec:framework-intervention}

Let $x_i$ contain the system prompt, tool description, and problem, and let $r_i$ be the model's stored with-tool reasoning prefix ending immediately before its original think-close boundary. A splice condition appends a sentence $\nu$ to this fixed prefix and regenerates only the continuation:
\begin{equation}
  p_i^{\nu}=x_i\,\Vert\,r_i\,\Vert\,\nu,
  \label{eq:splice-prefix}
\end{equation}
where $\Vert$ denotes token-sequence concatenation. The primary confidence and doubt interventions are
\begin{align}
  \nudgepos &: \quad \text{``I feel confident in my answer.''} \nonumber\\
  \nudgeneg &: \quad \text{``I do not feel confident in my answer.''}.
  \label{eq:nudge-sentences}
\end{align}
A neutral sentence, $\nudgeneutral$: ``This is the answer I arrived at.'', and a sentence-less splice, $\none$, serve as controls and are analyzed separately from the primary contrast. Appendix~\ref{app:anatomy} gives the exact splice prefixes, sentence sets, and family-specific think formats; Appendix~\ref{app:control-exp} lists the component each supporting experiment varies.

In the primary \emph{open-continuation} condition, the splice does not insert a think-close token after the sentence. The model may therefore absorb, reject, qualify, or otherwise reason about the signal before committing to an action. Figure~\ref{fig:prompt-anatomy} shows the full prompt anatomy. Within a paired comparison, the problem, system prompt, tool, reasoning prefix, splice boundary, and decoding configuration are identical; only the inserted sentence differs. The comparison therefore identifies the effect of replacing confidence with doubt at that fixed boundary.

\paragraph{Sensitivity.}
\label{sec:framework-sensitivity}

Let $\mathcal{I}_S$, called the retained set, contain problems with completed confidence and doubt continuations, and let $n_S=|\mathcal{I}_S|$. The paired delegation sensitivity is
\begin{equation}
  S = \frac{1}{n_S}\sum_{i\in\mathcal{I}_S}
  \left(
    \ind[A_i^{-}=\tool]-\ind[A_i^{+}=\tool]
  \right).
  \label{eq:sensitivity}
\end{equation}
Equivalently, $S$ is the delegation rate under doubt minus the delegation rate under confidence on the same retained problems. Positive $S$ means that doubt shifts behavior toward delegation relative to confidence; its magnitude measures sensitivity in percentage points. A completed \other{} action is treated as a non-tool outcome for this estimator. Because the two arms share the same prefix and regeneration procedure, effects common to both arms cancel from the contrast.

\paragraph{Targeting.}
\label{sec:framework-targeting}

Define the competence-appropriate action
\begin{equation}
  a^\star(s_i)=
  \begin{cases}
    \self, & s_i=1,\\
    \tool, & s_i=0.
  \end{cases}
  \label{eq:appropriate-action}
\end{equation}
Thus a model that succeeds unaided should answer for itself, whereas a model that fails unaided should delegate. This definition evaluates the delegation decision.

For each intervention polarity $\nu\in\{-,+\}$, let $\mathcal{I}_\nu$ contain problems with known $s_i$ and readable binary actions $A_i^B,A_i^\nu\in\{\self,\tool\}$. Define a flip indicator $f_{i\nu}$, a well-targeted-flip indicator $h_{i\nu}$, and the combined well-targeted share $\mathrm{WT}$ as
\begin{equation}
  \begin{aligned}
    f_{i\nu} &= \ind[A_i^B\neq A_i^\nu],\\
    h_{i\nu} &= f_{i\nu}\,\ind[A_i^\nu=a^\star(s_i)],
  \end{aligned}
  \qquad\qquad
  \mathrm{WT}=
  \frac{
    \sum_{\nu\in\{-,+\}}
    \sum_{i\in\mathcal{I}_\nu} h_{i\nu}
  }{
    \sum_{\nu\in\{-,+\}}
    \sum_{i\in\mathcal{I}_\nu} f_{i\nu}
  }.
  \label{eq:well-targeted-share}
\end{equation}

The denominator counts flip events rather than unique problems, so a problem that changes action under both confidence and doubt contributes twice. Both flip directions are retained and judged by their actions.
This keeps sensitivity and targeting conceptually separate. Transitions involving \other{} are excluded from both the numerator and denominator.

This definition is deliberately permissive and agnostic to the intended semantics of the inserted sentence. It asks only whether the \emph{action} of an induced flip agrees with observed unaided competence; it does not additionally require doubt to cause a \selftotool{} transition or confidence to cause a \tooltoself{} transition. Stricter alternatives could enforce those polarity--direction pairings, require the delegated tool call to succeed, or require the final answer to improve. Such definitions could classify additional flips as failures and therefore produce less favorable estimates. We use the action-based definition so that targeting does not inherit an assumed meaning for the confidence and doubt sentences: sentence semantics determine the sensitivity contrast, while competence determines whether the resulting action is useful in the limited sense measured here.

We calculate a random-selection baseline for $\mathrm{WT}$, which we call $\mathrm{WT}_{\mathrm{rand}}$, and define the lift $L = \mathrm{WT} - \mathrm{WT}_{\mathrm{rand}}$. This lift measures whether the intervention's actual action changes select competence-appropriate problems better than random allocation under the same action opportunities and flip volume. See Appendix~\ref{app:framework-random-reference} for details of how we calculate $\mathrm{WT}_{\mathrm{rand}}$.

\paragraph{Interpreting $S$ and $L$.}
Together, the two dimensions place each model--task pair in a behavioral profile. $S=100\pp$ means the model always answers unaided after the confidence sentence and always delegates after the doubt sentence; $S=0$ means the sentence does not change delegation on average, and $S<0$ that the model moves against it, delegating more after the confidence sentence than after the doubt sentence. $L=0$ means induced flips land where random selection from the same opportunities would place them, $L>0$ that they favor problems on which the new action matches unaided competence, and $L<0$ that they favor the wrong problems. The dimensions trade off at the extreme: a model that follows the signal completely flips every available problem, so its flips coincide with the random reference and $L=0$ by construction. A competence-aware model instead combines moderate $S$ with large $L$, following doubt on problems it fails and confidence on problems it solves while overriding the signal elsewhere. We therefore distinguish \emph{inert} ($S\approx 0$), \emph{contrarian} ($S<0$), \emph{compliant} (large $S$, $L\approx0$), \emph{selective} ($L>0$) and \emph{misdirected} ($L<0$) profiles.

\section{Experimental Setup}
\label{sec:experimental-setup}

\paragraph{Models.}
\label{sec:setup-models}
Our primary cohort contains nine open-weight reasoning models: Qwen3 (4B, 8B, 14B, 32B; FP8), Gemma4 (E2B, E4B, 31B; W4A16), and GLM-Z1 (9B, 32B; BNB4). Family, scale, and quantization are not factorially crossed, so comparisons among them are descriptive. We also evaluate DeepSeek-V4-Pro and MiniMax-M3 through provider APIs. Their continuation semantics, revisions, and precision are less auditable than local decoding, so we report these \emph{provider-served} models separately and do not pool the cohorts.

\paragraph{Tasks and tools.}
\label{sec:setup-tasks}
We use two multi-hop question-answering tasks with verifiable answers, chosen after preliminary scouting so that no model is near ceiling or floor unaided, since targeting needs both successes and failures. Unaided accuracy of the open-weight models spans $\unaidedAcc{musique_2hop}{open-weight-min}$--$\unaidedAcc{musique_2hop}{open-weight-max}\%$ on MuSiQue and $\unaidedAcc{strategyqa}{open-weight-min}$--$\unaidedAcc{strategyqa}{open-weight-max}\%$ on StrategyQA (medians $\unaidedAcc{musique_2hop}{open-weight-median}\%$ and $\unaidedAcc{strategyqa}{open-weight-median}\%$; Appendix~\ref{app:results-detailed}). StrategyQA consists of implicit multi-step yes/no questions \citep{geva2021strategyqa}. Its scoped tool permits a search for external evidence needed to answer the question. MuSiQue contains compositional multi-hop questions \citep{trivedi2022musique}; we use its two-hop reading-comprehension setting, supply the associated passages in the prompt, and offer a retrieval helper that selects the most relevant supplied passage. The two tasks therefore contrast open-web fact finding with bounded retrieval from an available context.

\paragraph{Prompts and conditions.}
\label{sec:setup-conditions}
Each problem has an unaided and a tool-aware prompt; the latter specifies the tool and the answer or tool-call format. All splice conditions (Section~\ref{sec:framework-intervention}) reuse the with-tool base's reasoning prefix and leave the reasoning block open after insertion. In the supporting \emph{free perfect oracle} condition, \texttt{ask\_expert} replaces the scoped tool; the prompt describes it as returning a correct worked answer (Appendix~\ref{sec:framework-diagnostics}). Appendix~\ref{app:control-exp} describes the supporting experiments and gives a complete example prompt (Figure~\ref{fig:prompt-example}).

\paragraph{Generation and sample sizes.}
\label{sec:setup-generation}
Open-weight experiments use greedy decoding with nominal $n=1000$ problems per model--task pair; provider-served experiments use $n=250$, and the sampled-decoding and repeated-label checks $n=100$. Retained sample sizes differ by estimator (Section~\ref{sec:framework-setting}); Appendix~\ref{app:stat-analysis} describes the statistical analysis.

\section{Results}
\label{sec:results}

We find that confidence language consistently changes whether models answer or delegate, but the problems on which behavior changes are only weakly related to unaided competence. Across the 18 open-weight model--task experiments, the median confidence-to-doubt delegation swing is $20.6\pp$. The median well-targeted share is $42\%$, only $+2\pp$ above the matched random-selection reference. The intervention therefore exposes a strong control surface, but not yet a reliable delegation policy.

\paragraph{Confidence and doubt consistently steer delegation.}
\label{sec:results-sensitivity}

The primary confidence--doubt contrast has the predicted sign in every open-weight model--task experiment. Inserting doubt increases tool use relative to inserting confidence, despite the fact that both continuations begin from the same problem, prompt, and model-generated reasoning prefix. The median swing of $20.6\pp$, pooled across both tasks, is therefore not attributable to different problems or different preceding trajectories across arms. It measures the behavioral effect of changing the valence of one sentence at the splice boundary.

This directional response is not confined to one family or task. It appears across Qwen, Gemma, and GLM models on both StrategyQA and MuSiQue. The provider-served DeepSeek and MiniMax models show the same direction, with the largest confidence-to-doubt swings exceeding $60\pp$. The effect therefore extends to frontier-scale models served through provider APIs. Appendix~\ref{app:frontier} records their serving routes and pooling.

The sampled-decoding study tests whether the sign is an artifact of greedy continuation. Across three models, two tasks, and three decoding seeds ($n=100$ problems per experiment; temperature 1.0, top-$p$ 0.95), all 18 seed-specific swings remain positive. Each of the six problem-clustered pooled intervals also excludes zero. Sampling changes the magnitude of the response, but not its direction in this subset.

\newcommand{\resultPoint}[3]{\newEvidence{#1}{#2}{#3-point}}
\newcommand{\liftMark}[2]{\ifdim\newEvidence{#1}{#2}{lift-low}pt>0pt\rlap{$^\dagger$}\else\ifdim\newEvidence{#1}{#2}{lift-high}pt<0pt\rlap{$^\dagger$}\fi\fi }
\newcommand{\resultCondensedRow}[2]{#2 & \resultPoint{musique_2hop}{#1}{S} & \resultPoint{musique_2hop}{#1}{wt} & \resultPoint{musique_2hop}{#1}{lift}\liftMark{musique_2hop}{#1} & \resultPoint{strategyqa}{#1}{S} & \resultPoint{strategyqa}{#1}{wt} & \resultPoint{strategyqa}{#1}{lift}\liftMark{strategyqa}{#1} \\}
\newcommand{\resultMedianRow}{\textit{Median (open-weight)} & \resultPoint{musique_2hop}{open-weight-median}{S} & \resultPoint{musique_2hop}{open-weight-median}{wt} & \resultPoint{musique_2hop}{open-weight-median}{lift} & \resultPoint{strategyqa}{open-weight-median}{S} & \resultPoint{strategyqa}{open-weight-median}{wt} & \resultPoint{strategyqa}{open-weight-median}{lift} \\}

\begin{table}[t]
\centering
\caption{Confidence--doubt delegation swing ($S$), well-targeted flip share ($\mathrm{WT}$), and targeting lift ($L$) across both datasets. Nominal $n=1000$ for open-weight models and $n=250$ for provider-served models. Point estimates are reported here; $^\dagger$ marks a lift whose 95\% problem-bootstrap interval excludes zero, and the median row is taken per task over the nine open-weight models. Full 95\% pointwise problem-bootstrap intervals, retained sample sizes ($n_S/n_{\mathrm{WT}}$), flip event counts (H/F), and random baselines appear in Appendix~\ref{app:results-table}.}
\label{tab:main-results}
\small
\setlength{\tabcolsep}{6pt}
\renewcommand{\arraystretch}{1.02}
\begin{tabular}{l rrr rrr}
\toprule
& \multicolumn{3}{c}{\textbf{MuSiQue 2-hop}} & \multicolumn{3}{c}{\textbf{StrategyQA}} \\
\cmidrule(lr){2-4} \cmidrule(lr){5-7}
Model & $S$ (pp) & WT (\%) & $L$ (pp) & $S$ (pp) & WT (\%) & $L$ (pp) \\
\midrule
\multicolumn{7}{l}{\textbf{Open-weight models} ($n=1000$)} \\[1pt]
\resultCondensedRow{qwen3-4b-fp8}{Qwen3-4B}
\resultCondensedRow{qwen3-8b-fp8}{Qwen3-8B}
\resultCondensedRow{qwen3-14b-fp8}{Qwen3-14B}
\resultCondensedRow{qwen3-32b-fp8}{Qwen3-32B}
\resultCondensedRow{gemma4-e2b-w4a16}{Gemma4-E2B}
\resultCondensedRow{gemma4-e4b-w4a16}{Gemma4-E4B}
\resultCondensedRow{gemma4-31b-w4a16}{Gemma4-31B}
\resultCondensedRow{glm-z1-9b-bnb4}{GLM-Z1-9B}
\resultCondensedRow{glm-z1-32b-bnb4}{GLM-Z1-32B}
\cmidrule(lr){1-7}
\resultMedianRow
\specialrule{0.8pt}{2pt}{3pt}
\multicolumn{7}{l}{\textbf{Provider-served models} ($n=250$)} \\[1pt]
\resultCondensedRow{deepseek-v4-pro}{DeepSeek-V4-Pro}
\resultCondensedRow{minimax-m3}{MiniMax-M3}
\bottomrule
\end{tabular}
\end{table}

\paragraph{Behavioral movement is weakly targeted.}
\label{sec:results-targeting}

Sensitivity does not imply that the model changes its action on the right problems. Pooled across the open-weight experiments on both tasks, the median well-targeted share of induced flips is $42\%$.
Relative to $\mathrm{WT}_{\mathrm{rand}}$, the median targeting lift is only $+2\pp$. Appendix~\ref{app:results-detailed} also shows how helpful and harmful transitions compose that ratio in three illustrative models.

\begin{figure*}[t]
\centering
\includegraphics[width=0.9\linewidth]{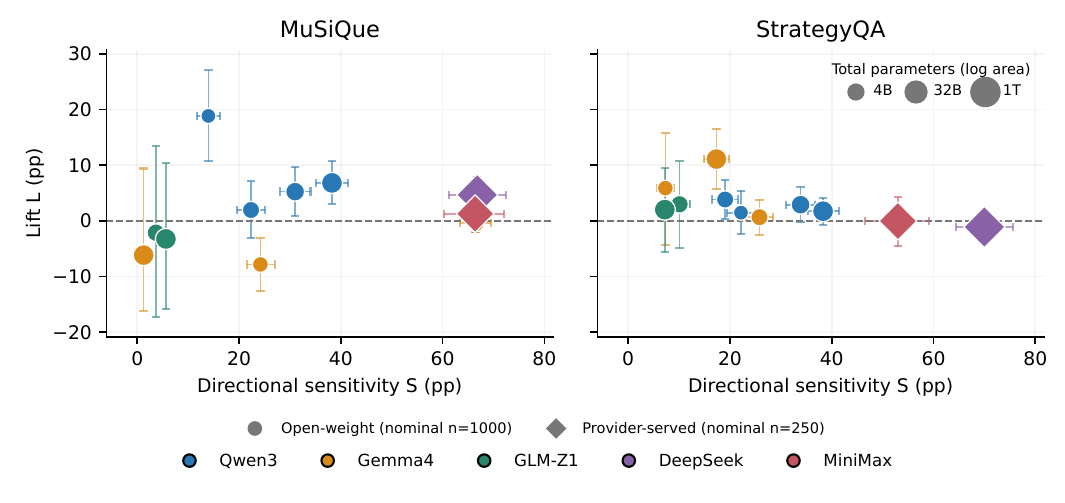}
\caption{Directional sensitivity versus targeting relative to random selection for the 22 model--task experiments from Table~\ref{tab:main-results}. Horizontal and vertical whiskers show marginal 95\% problem-bootstrap intervals for $S$ and $L$, respectively; they are not a joint confidence region. Color denotes model family, distinct marker shapes distinguish the two cohort scopes, and marker area grows with the logarithm of total parameter count. }
\label{fig:movement-value}
\end{figure*}

The small lift shows that the intervention's selection of problems is only slightly better than random allocation. Large values of $S$ can therefore coexist with weak or negative $L$. Figure~\ref{fig:movement-value} shows this separation directly by plotting sensitivity against targeting lift for every model--task experiment. In practical terms, the model often reacts to the semantic direction of the sentence without reliably connecting that reaction to whether its unaided answer would be correct. Most induced flips are \selftotool{} transitions: 72\% pooled, and a majority in 17 of 18 open-weight experiments. This separation between \emph{movement} and \emph{selection} is the central empirical result.

The experiments occupy all but the \emph{contrarian} profile ($S>0$ throughout; Figure~\ref{fig:movement-value}). GLM-Z1 ($S=3.7$--$10.1\pp$) and Gemma4-31B on MuSiQue ($1.3\pp$, at a $0.7\%$ base delegation rate) are nearly \emph{inert}. The largest responses are \emph{compliant}: Gemma4-E4B on MuSiQue and the provider-served models move by $53$--$70\pp$, yet their lifts lie between $-1.1$ and $+4.6\pp$. The clearest \emph{selective} profiles are Qwen3 on MuSiQue (4B, 14B, 32B; $L=+5.2$ to $+18.8\pp$) and Gemma4-31B on StrategyQA ($+11.1\pp$), while Gemma4-E2B on MuSiQue is \emph{misdirected} ($L=-7.9\pp$). Profiles depend on the task and its scoped tool as much as on the model: $L$ is positive in every StrategyQA experiment but negative in five of nine on MuSiQue, and across experiments $S$ does not predict $L$ (Spearman $\rho=0.13$). The models' own routing already carries competence information: without any inserted sentence, the with-tool base delegates more on unaided failures than on successes in 17 of 18 open-weight experiments (median $\Gamma=P(A^B{=}\tool\mid s{=}0)-P(A^B{=}\tool\mid s{=}1)=+5.6\pp$; Appendix~\ref{app:natural-oracle}). The induced flips make little use of it.

Even a free, guaranteed-correct oracle leaves substantial self-reliance: models still answer directly on $19$--$96\%$ of problems, a spread driven mainly by task ($64$--$96\%$ on MuSiQue; mostly $19$--$44\%$ on StrategyQA; Appendix~\ref{app:natural-oracle}). Appendix~\ref{app:retained-self-correctness} reports correctness transitions among the retained self-answers.

Further control experiments (Appendix~\ref{app:controls}) show that the main effect is not tied to a single superficial implementation choice.
Cutting and regenerating without a sentence is nearly inert (median $|\,\none - A^B| = 0.3\pp$, at most $3.6\pp$; Table~\ref{tab:placebo-arms}); the neutral sentence is usually inert relative to $\none$ (median $-0.9\pp$) but not always ($-31.4\pp$ for Qwen3-32B on StrategyQA), which is why $S$ contrasts confidence with doubt; first-person attribution moves delegation at least as much as expert attribution in 16 of 18 experiments (Table~\ref{tab:voice}); and sealing the think block right after the sentence keeps $S>0$ in all 18 while changing its magnitude (Appendix~\ref{app:closed}).

\section{Discussion and Limitations}
\label{sec:discussion}

\paragraph{What Nudgeability establishes.}

The central result is a separation between \emph{behavioral sensitivity} and \emph{competence-aware control}.
The consistency and magnitude of the Nudgeability scores across model families, datasets, and sizes establish confidence language inside reasoning as a potential control surface for the model's subsequent action.

That control surface does not, by itself, constitute a useful delegation policy. Ideally, a model would follow the nudge only where it agrees with its unaided competence, delegating under doubt on problems it fails and overriding doubt on problems it solves, and vice versa, so that every induced flip is well targeted ($\mathrm{WT}=100\%$). Instead, the induced action changes are only weakly concentrated on problems where the model's unaided behavior indicates that help is needed. A model can therefore act on the meaning of ``I do not feel confident'', delegating more, with little regard for whether it would actually fail the current problem.

Most models do react to it without being (post-)trained explicitly, but the models are unable to use this signal in a competence appropriate manner.
A future uncertainty estimator could be highly predictive yet operationally ineffective if the reasoning model ignores the representation through which it is supplied.
An endogenous reflective mechanism must join these components: a grounded signal, a usable representation, and a downstream response that is both sensitive and well targeted.

Our natural-language intervention is useful partly because it is a common interface across the evaluated models.
This makes Nudgeability a post-training-free diagnostic property of the model.
Confidence tokens, continuous feedback, or hidden-state interventions can be evaluated on the same footing, and may yield different profiles.
Our framework is intended to make those representations comparable at the level of subsequent behavior.

\paragraph{Implications for tool-using systems.}
\label{sec:implications}

For agent design, the results argue against evaluating a delegation mechanism only by how much it changes tool-use rates. An intervention that strongly raises delegation can waste latency and cost on problems the model already solves; one that strongly suppresses delegation can preserve confident errors. Reporting sensitivity together with targeting lift makes this failure visible. In particular, a model with high residual self-reliance and high conditional overconfidence under the oracle control should not be the sole gatekeeper of its access to stronger tools. Appendix~\ref{app:results-detailed} reports the results of the experiments.

A practical analogue is a local model deciding whether to defer to a larger remote model. The remote model is neither free nor guaranteed correct, so the oracle condition is an upper bound rather than a deployable policy. A real router must trade expected accuracy gain against latency, monetary cost, privacy, and failure risk. Nudgeability measures one prerequisite for such routing---whether a supplied reflective signal can change the action---while $L$ tests whether that change is directed toward cases that need help.

The results also identify the reasoning trace as an influential intervention surface. ``Privileged'' here should be understood as \emph{privileged influence}, not privileged epistemic access: editing a short span inside the reasoning region can exert a large causal effect on a later action. This does not show that the displayed trace faithfully represents the model's internal computation, nor that naturally generated confidence statements would have the same effect. It does motivate treating reasoning-channel interventions as part of behavioral and safety evaluation rather than as inert explanatory text.

A limitation of our work is that the primary observation we make is whether the model calls the tool, not whether executing that call would improve the final answer.
Scoped tools can fail, return irrelevant evidence, or cost more than the expected benefit.
$\mathrm{WT}$ and $L$ score the direction of the decision against unaided competence; they do not estimate downstream reward. Only the simulated oracle fixes the value of delegation by construction.

Another limitation is that the evaluation covers two question-answering tasks, three open-weight model families, and two provider-served models. Model family, scale, quantization, and serving stack are not factorially crossed, so apparent size or family patterns are descriptive. MuSiQue's bounded retrieval and StrategyQA's web-oriented search do not span the range of tools used by deployed agents.

\paragraph{From an injected signal to endogenous reflection.}

The natural next step is to close the loop between signal quality and behavioral response. An uncertainty estimate derived from sampled outputs, a semantic-entropy probe, a hidden-state predictor, or an activation-level monitor could be mapped into the same intervention interface and evaluated using $S$, WT, and $L$. Extending semantic uncertainty probes to reasoning models \citep{kossen2024semantic} would test how much of the available control surface can be converted into competence-aware delegation. Activation monitoring and control provide a complementary route that need not pass through natural-language reports \citep{li2025metacognitive}.

\section{Conclusion}
\label{sec:conclusion}

A single sentence expressing confidence or doubt inside a fixed reasoning trajectory can substantially change whether a reasoning model answers directly or delegates to a tool. Across the open-weight experiments, the median confidence-to-doubt swing is $20.6\pp$, and the larger provider-served models exhibit swings of $53$--$70\pp$. Yet only a median $42\%$ of induced flips are well targeted, amounting to a median lift of just $+2\pp$ over matched random selection. The models are therefore reliably steerable through confidence language, but only weakly selective about where that steering should apply.
A practical reflective system ultimately needs both properties: an uncertainty signal grounded in correctness and an action response that uses that signal selectively.

With a paired, post-training-free intervention, Nudgeability separates sensitivity from targeting: sensitivity measures whether the signal moves behavior, and targeting measures whether it moves behavior on the right problems. This distinction provides a simple test for future self-reflection mechanisms. Improving how a model produces or represents uncertainty is not sufficient by itself; the model must also translate that signal into competence-appropriate action.

\section*{Reproducibility Statement}
All models, prompts, decoding settings, and estimators are specified in the paper, in Section~\ref{sec:setup-models}, and the provider-served models with their serving routes in Appendix~\ref{app:frontier}.
Problems are sampled from MuSiQue and StrategyQA with a fixed seed (Section~\ref{sec:setup-generation}). Figure~\ref{fig:prompt-example} shows a complete prompt, and Appendix~\ref{app:anatomy} gives the exact splice prefixes and inserted sentences. Primary runs use greedy decoding; the sampled-decoding check uses temperature 1.0, top-$p=0.95$, and three decoding seeds (Appendix~\ref{app:stochastic}). All statistics are deterministic post-processing of those records. The estimators are defined in Section~\ref{sec:framework} and Appendix~\ref{app:framework-random-reference}, and the bootstrap procedure in Appendices~\ref{app:framework-random-reference} and~\ref{app:stat-analysis}. The code will be released upon publication.

\section*{AI Use Disclosure}
In this work, we used generative AI tools (Claude, OpenAI, DeepSeek, and GLM models) for the following purposes. \emph{Writing and drafting:} to polish and condense prose, draft and restructure sections of the manuscript, and prepare LaTeX for tables and figures from stored results. \emph{Retrieval and discovery:} to locate and summarize related work; the authors checked every cited work against its original source. \emph{Research ideation and execution:} to implement and debug the experiment code and data processing. The authors reviewed all AI-assisted content and take full responsibility for the paper.

\bibliographystyle{plainnat}

\clearpage
\appendix

\section{Related Work}
\label{app:related-work}

Our work lies at the intersection of self-reflection, uncertainty estimation, reasoning-trace interventions, and selective tool use. We organize these literatures using the distinction developed in Section~\ref{sec:introduction}: the source of a reflective signal, its representation to the model, and its effect on a subsequent action. Existing methods often address more than one of these questions simultaneously. Our experiment deliberately fixes the first two in order to identify the third.

\paragraph{Verbal reflection and sources of uncertainty.}
Reflexion converts task feedback into verbal reflections that can guide later decisions, while Self-Refine iteratively generates feedback and uses it to revise an earlier output \citep{shinn2023reflexion,madaan2023selfrefine}. In both cases, a model-produced textual signal is returned to the model as context. A separate line of work asks whether models can estimate the correctness of their own answers. Such estimates can be elicited from verbal or probabilistic reports \citep{kadavath2022know}, computed from the semantic diversity of sampled answers \citep{farquhar2024semantic}, or predicted from hidden representations \citep{kossen2024semantic,ghasemabadi2025gnosis}. These approaches investigate where useful information about correctness can be found and how accurately it can be recovered. Nudgeability instead asks what happens after a candidate signal has been supplied: whether the model's next action responds to it, and whether that response is aligned with unaided competence.

This distinction also limits the introspective claim supported by our study. Under the criterion proposed by \citet{song2025privileged}, evidence of introspection requires privileged access to self-information rather than merely behavior that appears self-aware. Because our confidence and doubt sentences are supplied by the experimenter, their effect cannot establish that the model extracted, estimated, or truthfully reported its own uncertainty. The intervention tests control through reflection-like language, not the provenance or faithfulness of that language.

\paragraph{Representing reflective signals inside computation.}
The same uncertainty estimate can be exposed to a model through substantially different interfaces. Learned confidence tokens make uncertainty available in a discrete form designed for routing \citep{chuang2025confidence}, while explicit reasoning-time markers train models to express uncertainty within their generated traces \citep{guo2026uncertainty}. Soft Thinking instead feeds a probability-weighted output-distribution representation back into ongoing reasoning instead of tokens themselves \citep{zhang2025softthinking}. Activation-neurofeedback experiments intervene on internal directions and test whether models can monitor and modulate those activations \citep{li2025metacognitive}. These methods couple particular signal sources to particular representations. Our natural-language sentence is one deliberately simple representation: it requires no fine-tuning and can be inserted at a known point in an existing trajectory. We do not claim that first-person prose is uniquely effective. Rather, the sensitivity and targeting dimensions of Nudgeability provide questions that can also be asked of future confidence tokens, hidden-state readouts, or continuous feedback mechanisms.

\paragraph{Reasoning-trace interventions and faithfulness.}
Thinking Intervention demonstrates that inserting or revising tokens inside a reasoning trace can control a reasoning model's output \citep{wu2025thinking}; dynamic early-exit methods likewise alter behavior by changing where reasoning terminates \citep{yang2025earlyexit}. Work on chain-of-thought faithfulness edits, truncates, or perturbs reasoning to measure whether final answers depend on the displayed trace \citep{lanham2023faithfulness}, while bias studies show that plausible explanations may omit factors that affected the answer \citep{turpin2023unfaithful}. Self- and cross-model counterarguments further show that answer stability can depend on the apparent source of an argument \citep{nikeghbal2026flips}.

We share the interventional premise that a displayed reasoning trace should not be treated as a faithful causal account without testing it. Our focus differs in four respects. We vary one semantic feature---expressed confidence---at the same boundary of a fixed prefix; measure a discrete agentic decision rather than answer content alone; pair counterfactual arms on the same problem and trajectory; and evaluate whether induced action changes occur on problems where delegation is warranted by unaided performance. Thus the contribution is not the observation that reasoning text can steer outputs, but a framework for paired evaluation of how strongly and how selectively it steers delegation.

\paragraph{Tool use, routing, and abstention.}
ReAct integrates reasoning with actions such as tool calls, making the decision to seek external information part of the model's generated trajectory \citep{yao2023react}. \citet{wu2026call} distinguish the necessity, utility, and affordability of tool calls and optimize the call/no-call decision using hidden-state controllers. That work asks how to construct an effective routing policy; we instead hold the decision inside the reasoning model and intervene on a signal that may influence it. The two perspectives are complementary: a reliable uncertainty estimator or controller supplies a possible reflective signal, whereas Nudgeability measures whether a model changes its own delegation behavior when such a signal is presented.

Delegation is also related to abstention. Prior work evaluates whether prompting, confidence thresholds, or chain-of-thought techniques help language models refrain from answering when they are likely to be wrong \citep{madhusudhan2025abstention}, and \citet{wen2025abstention} survey abstention objectives and evaluation methods. Abstention ends or defers an answer, whereas a tool call may continue the same attempt by acquiring information or computation. Nevertheless, both require distinguishing cases where an unaided answer should not be trusted. Our targeting measure makes that connection explicit by testing whether intervention-induced delegation increases on problems the model fails unaided and decreases on those it solves.

\section{Control Experiments}
\label{app:control-exp}

\begin{figure*}[!ht]
  \centering
  \includegraphics[width=\textwidth]{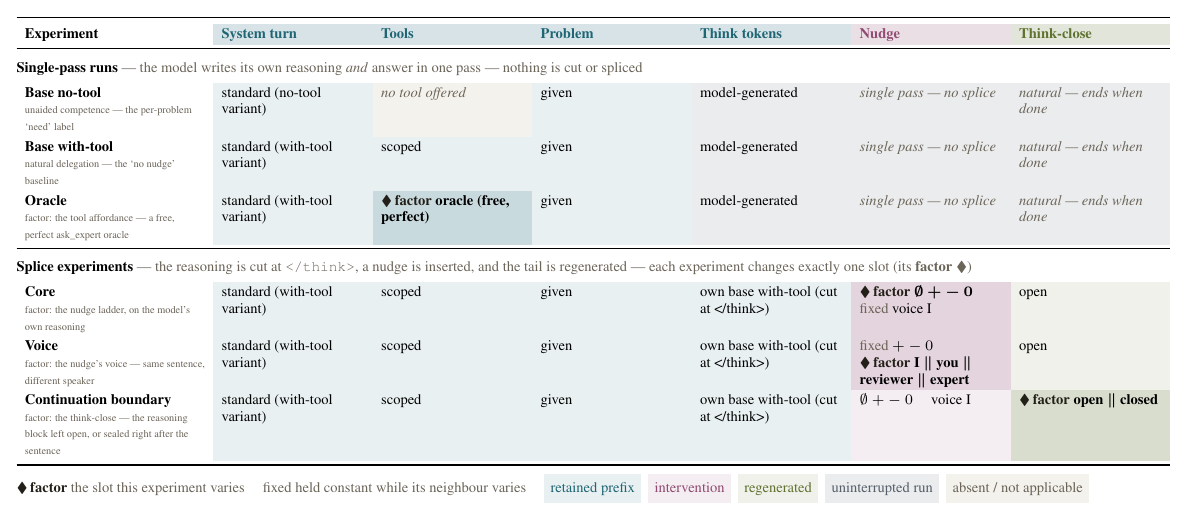}
  \caption{How the six prompt components are populated across single-pass and splice experiments. The highlighted factor in each row identifies the component varied by that experiment; the remaining cells show what is held fixed or inherited from a base run.}
  \label{fig:prompt-anatomy-experiments}
\end{figure*}

\begin{figure*}[!ht]
  \centering
  \includegraphics[width=\textwidth,height=0.84\textheight,keepaspectratio]{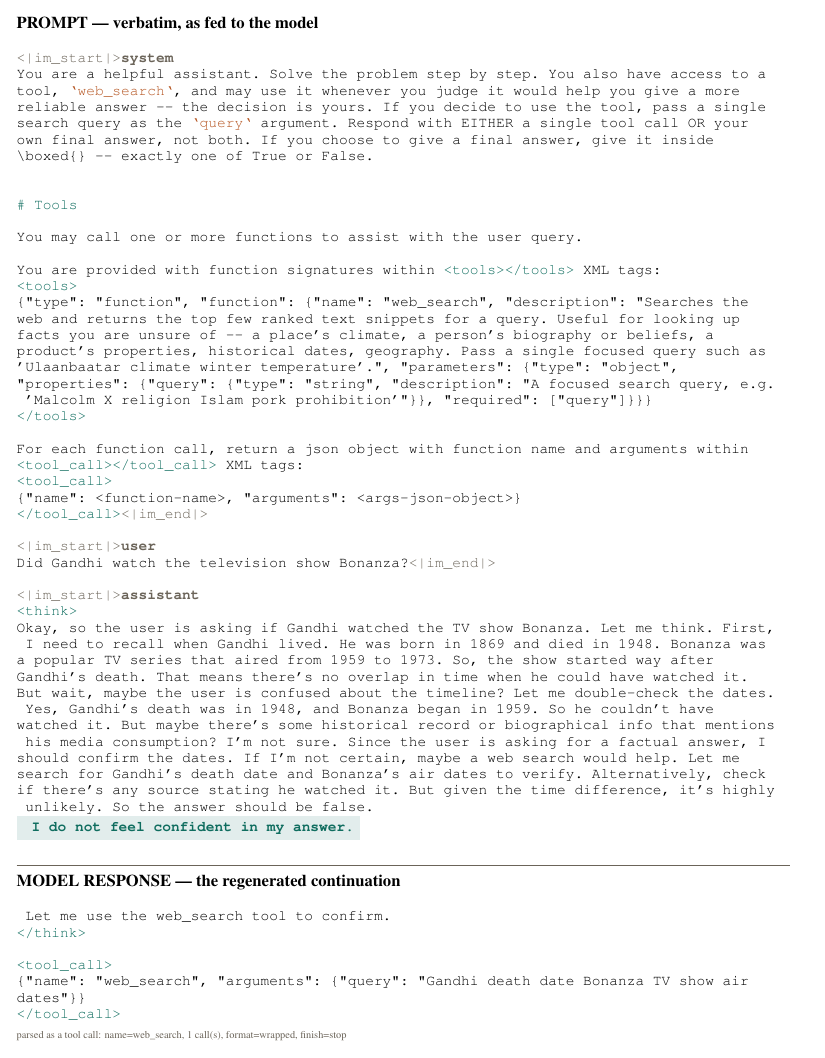}
  \caption{A complete spliced prompt and the model's response --- Qwen3 family, StrategyQA dataset, scoped \texttt{web\_search} tool, with tool reasoning cut at \texttt{</think>}, a first-person doubt sentence inserted, think block left open. Below the rule is the model's regenerated continuation, which closes the think block and calls the tool.}
  \label{fig:prompt-example}
\end{figure*}

Supporting experiments change one factor at a time:

\begin{enumerate}[leftmargin=*,itemsep=2pt,topsep=3pt]
\item \textbf{Voice.} The semantic confidence content is attributed to the model itself, a second-person observer, a reviewer, or an expert. This tests whether self-attribution strengthens the response.
  \item \textbf{Oracle.} The scoped tool is replaced with the free perfect oracle to separate unwillingness to delegate from the effort or uncertainty involved in formulating a tool request.
  \item \textbf{Continuation boundary.} A sealed condition inserts the model-specific think-close marker immediately after the nudge. Comparing it with the open condition tests whether further deliberation absorbs or amplifies the intervention, while recognizing that sealing also changes the continuation boundary.
\end{enumerate}
The confidence--doubt comparison remains the primary estimand. The sentence-less, neutral, voice, oracle, and sealed conditions are controls or supporting analyses rather than additional components of the headline effect.
Figure~\ref{fig:prompt-anatomy-experiments} maps each base and splice experiment to the six prompt components and marks the factor varied by each control.
Figure~\ref{fig:prompt-example} shows a complete example prompt.

\section{Intervention Anatomy}
\label{app:anatomy}

For a core splice, the prompt is the original with-tool prompt plus a partial assistant continuation, denoted by $x_i$. Let $r_i$ be the exact reasoning text from the model's with-tool base. The four open prefixes are
\begin{align*}
  p_i^{\none} &= x_i \Vert r_i,\\
  p_i^{+} &= x_i \Vert r_i \Vert \text{``I feel confident in my answer.''},\\
  p_i^{-} &= x_i \Vert r_i \Vert \text{``I do not feel confident in my answer.''},\\
  p_i^{0} &= x_i \Vert r_i \Vert \text{``This is the answer I arrived at.''}.
\end{align*}
The model then continues the same assistant turn. The closed sanity check appends the model-specific think-end marker after the sentence. Model adapters preserve each family's native dialect: Qwen3 uses \texttt{<think>... </think>}; Gemma4 uses channel markers; and GLM-Z1 uses a template-primed think opener followed by a generated close. Tool-call and answer parsing are likewise adapter-specific.

The four voice sets preserve answer-confidence content while changing attribution:
\begin{center}
\small
\resizebox{\linewidth}{!}{\begin{tabular}{lll}
\toprule
Voice & Positive & Negative \\
\midrule
First person & I feel confident in my answer. & I do not feel confident in my answer. \\
Second person & You seem confident in your answer. & You do not seem confident in your answer. \\
Reviewer & A reviewer is confident in this answer. & A reviewer is not confident in this answer. \\
Expert & An expert is confident in this answer. & An expert is not confident in this answer. \\
\bottomrule
\end{tabular}}

\end{center}
The corresponding neutral sentences are, respectively, ``This is the answer I arrived at,'' ``This is the answer you arrived at,'' ``A reviewer has considered this answer,'' and ``An expert has considered this answer.'' The sentence-less splice $\none$ controls the cut/regeneration mechanism; it is distinct from the uninterrupted with-tool base used to identify induced flips.

\section{Supporting delegation diagnostics}
\label{sec:framework-diagnostics}

We use several auxiliary quantities to contextualize our findings.
First, the uninterrupted with-tool base provides a \emph{natural delegation targeting gap}
\begin{equation}
  \Gamma=
  P(A^B=\tool\mid s=0)-P(A^B=\tool\mid s=1).
  \label{eq:natural-targeting-gap}
\end{equation}
Positive $\Gamma$ means that, without an inserted sentence, the model delegates more often on problems it fails unaided than on problems it solves.

Tool availability can also change the answer without changing the action.
Let $y^B=1$ when the uninterrupted with-tool base answers for itself and that answer is correct.
Among paired base runs retained as \self{}, we measure \emph{retained-self degradation} and \emph{recovery}:
\begin{align}
  D_{\mathrm{self}} &= P(y^B=0\mid s=1, A^B=\self), \\
  R_{\mathrm{self}} &= P(y^B=1\mid s=0, A^B=\self).
  \label{eq:retained-self}
\end{align}
Both hold $s$ fixed as the unaided competence label and ask whether a fresh self-answer under the tool-aware prompt crosses the correctness boundary.
A problem is eligible only when the with-tool base commits to \self{} without truncation and both its answer and the no-tool answer are gradeable.
Neither quantity assigns correctness to a delegated row.
They are reported in Appendix~\ref{app:retained-self-correctness}.

We also perform experiments after replacing the scoped tool with a free perfect oracle called \texttt{ask\_expert}. This control removes search and request-formulation difficulty, allowing \emph{residual self-reliance}, $\rho$, to be measured when delegation is costless and guaranteed to provide a correct response.
\begin{align}
  \rho=P(A=\self\mid\text{free perfect oracle}), \\
  \omega=P(s=0\mid A=\self,\text{free perfect oracle}).
  \label{eq:oracle-self-reliance}
\end{align}
The quantity $\omega$, called \emph{conditional overconfidence}, is the fraction of retained self-answers that occur on problems the model failed unaided, despite access to a guaranteed-correct option. These diagnostics describe baseline behavior and supporting controls; $S$, $\mathrm{WT}$, and $L$ remain the primary Nudgeability measures.

\section{Random-Selection Reference}
\label{app:framework-random-reference}

The value $50\%$ is not, in general, a chance baseline for $\mathrm{WT}$. The pool of problems available to flip can contain unequal numbers of unaided successes and failures, and the two baseline actions offer different opportunities for a helpful action. We therefore compare observed targeting with a matched random-selection reference that preserves the intervention arm, baseline action, eligible problem pool, and observed number of flips, but selects which problems flip without using competence.

For polarity $\nu\in\{-,+\}$ and baseline action $b\in\{\self,\tool\}$, let $\bar b$ denote the opposite binary action and define
\begin{align}
  N_{\nu b}
    &= \sum_{i\in\mathcal{I}_\nu}\ind[A_i^B=b],
    \label{eq:eligible-pool}\\
  K_{\nu b}
    &= \sum_{i\in\mathcal{I}_\nu}
       \ind[A_i^B=b]\ind[a^\star(s_i)=\bar b],
    \label{eq:helpful-opportunities}\\
  F_{\nu b}
    &= \sum_{i\in\mathcal{I}_\nu}
       \ind[A_i^B=b]f_{i\nu}.
    \label{eq:observed-flips}
\end{align}
Here, $N_{\nu b}$ is the eligible stratum size, $K_{\nu b}$ is the number of problems for which a flip from $b$ would land on the competence-appropriate action, and $F_{\nu b}$ is the observed number of flips in that stratum. Uniformly allocating those $F_{\nu b}$ flips among the $N_{\nu b}$ eligible problems gives an expected $F_{\nu b}K_{\nu b}/N_{\nu b}$ well-targeted flips. Aggregating the four strata for each polarity yields
\begin{equation}
  \mathrm{WT}_{\mathrm{rand}}=
  \frac{
    \displaystyle\sum_{(\nu,b)\in\mathcal{S}}
    F_{\nu b}\,K_{\nu b}/N_{\nu b}
  }{
    \displaystyle\sum_{(\nu,b)\in\mathcal{S}} F_{\nu b}
  },
  \qquad
  L=\mathrm{WT}-\mathrm{WT}_{\mathrm{rand}}.
  \label{eq:targeting-lift}
\end{equation}
where $\mathcal{S}=\{(\nu,b):N_{\nu b}>0\}$. Empty strata are omitted, and both quantities are undefined when no flips occur. The raw $\mathrm{WT}$ describes where the intervention's actual action changes land. The lift $L$ asks whether those changes select competence-appropriate problems better than random allocation under the same action opportunities and flip volume. Neither quantity measures the downstream value of the tool itself.

\label{app:wtnull}

Within each pool, a uniformly sampled subset of $F_{\nu b}$ problems has a hypergeometric helpful count with expectation $F_{\nu b}K_{\nu b}/N_{\nu b}$, which gives Eq.~\eqref{eq:targeting-lift}. An empty pool contributes zero because it contains no flips; WT and lift are undefined if the total flip count is zero. All four pools are necessary: the measured WT includes counterdirectional flips as well as doubt-induced \selftotool{} and confidence-induced \tooltoself. A problem that flips in both polarities contributes two events. Each polarity retains its own pairwise-complete domain, rather than requiring both interventions to be available.

We resample whole problems, carrying their contributions to every pool together, and recompute observed WT, pool composition, reference and lift inside each bootstrap draw. Zero-flip problems stay in the resampling population. We use 20,000 draws and master seed 20260911 with deterministic experiment-specific seeds. Intervals are marginal, pointwise 95\% percentile intervals, not a randomization test or multiplicity-adjusted declaration of superiority. Undefined zero-flip draws are omitted and their counts are retained in the generated evidence artifact; pilot draw counts are not reused for the expanded experiments.

At fixed flip counts, the attainable helpful count in a pool lies between $\max(0,F_{va}-(N_{va}-K_{va}))$ and $\min(F_{va},K_{va})$. Summing these bounds and dividing by total flips gives allocation bounds, not uncertainty intervals or a ceiling on better policies with different flip volumes. High sensitivity can exhaust the pool of helpful opportunities even with selective allocation.

\section{Statistical Analysis}
\label{app:stat-analysis}

All primary comparisons are paired by problem and fixed reasoning prefix. We report $S$ in percentage points, $\mathrm{WT}$ as a percentage with its well-targeted/all-flip count, and targeting lift $L$ relative to the matched random-selection reference. Uncertainty intervals are obtained by nonparametric bootstrap resampling of problem identifiers within each model--task experiment; every replicate recomputes the relevant estimator, including its random flip denominator. We use 20,000 bootstrap replicates and report 95\% intervals. These intervals quantify variation over the sampled problems under the fixed model and decoding configuration; they do not capture variation across checkpoints, serving stacks, or decoding seeds.

For the paired confidence--doubt tool indicators, we additionally use McNemar's test on discordant problem pairs to assess directional action changes. Statistical tests are performed within model--task experiments. Cross-model and cross-task medians summarize the resulting experiment-level estimates rather than pooling problem rows across heterogeneous models or tasks.

To test whether the greedy-decoding result is directionally robust, we repeat a fixed subset containing Qwen3-8B, Gemma4-E4B, and GLM-Z1-9B on both tasks with temperature 1.0, top-$p=0.95$, and three decoding seeds. Positive and negative arms remain paired within each seed. The sampled-decoding study is a sensitivity analysis of direction and variability, not a second estimate pooled with the primary greedy campaign. Appendix~\ref{app:stochastic} gives its seeding scheme, realized generation counts, and pooled intervals.

\section{Detailed Experimental Results}
\label{app:results-detailed}

\subsection{Full Sensitivity and Targeting Results}
\label{app:results-table}

Table~\ref{tab:unaided-accuracy} reports the unaided accuracy that defines the competence labels $s_i$.
\begin{table}[htbp]
\centering\small
\caption{Unaided accuracy of the no-tool base, $P(s=1)$: correct answers over non-truncated no-tool generations (truncated generations have unknown competence and are excluded). Nominal $n=1000$ for open-weight models and $n=250$ for provider-served models.}
\label{tab:unaided-accuracy}
\begin{tabular}{lrrrr}
\toprule
& \multicolumn{2}{c}{MuSiQue 2-hop} & \multicolumn{2}{c}{StrategyQA} \\
\cmidrule(lr){2-3} \cmidrule(lr){4-5}
Model & correct/known & Acc.\ (\%) & correct/known & Acc.\ (\%) \\
\midrule
\multicolumn{5}{l}{\textbf{Open-weight models}} \\[1pt]
Qwen3-4B & 521/889 & 58.6 & 711/991 & 71.7 \\
Qwen3-8B & 597/939 & 63.6 & 763/999 & 76.4 \\
Qwen3-14B & 625/990 & 63.1 & 757/999 & 75.8 \\
Qwen3-32B & 658/992 & 66.3 & 816/1000 & 81.6 \\
Gemma4-E2B & 508/993 & 51.2 & 616/999 & 61.7 \\
Gemma4-E4B & 616/1000 & 61.6 & 743/1000 & 74.3 \\
Gemma4-31B & 807/975 & 82.8 & 818/985 & 83.0 \\
GLM-Z1-9B & 504/867 & 58.1 & 731/991 & 73.8 \\
GLM-Z1-32B & 565/857 & 65.9 & 797/1000 & 79.7 \\
\cmidrule(lr){1-5}
\textit{Median (open-weight)} & & 63.1 & & 75.8 \\
\specialrule{0.8pt}{2pt}{3pt}
\multicolumn{5}{l}{\textbf{Provider-served models}} \\[1pt]
DeepSeek-V4-Pro & 196/250 & 78.4 & 211/249 & 84.7 \\
MiniMax-M3 & 197/241 & 81.7 & 206/250 & 82.4 \\
\bottomrule
\end{tabular}
\end{table}
 
Table~\ref{tab:null-full} reports the primary open-continuation results for every experiment: 18 open-weight experiments (nine models on two tasks) at nominal $n=1000$ and four provider-served experiments (two models on two tasks) at $n=250$.
For each experiment, the table gives the confidence--doubt delegation swing $S$ (percentage points), the well-targeted flip share WT (percent), its random-selection reference $\mathrm{WT}_{\mathrm{rand}}$ (percent), and the targeting lift, WT minus $\mathrm{WT}_{\mathrm{rand}}$ (percentage points). $\mathrm{WT}_{\mathrm{rand}}$ is the WT expected if the observed number of flips were drawn at random from the observed polarity-specific baseline-action opportunity pools (Appendix~\ref{app:wtnull}). H/F gives helpful/all flip events, including counterdirectional flips.
The two sample sizes describe different populations. $n_S$ is the positive/negative-paired population behind $S$; $n_{\mathrm{WT}}$ is the baseline/unaided resampling population behind WT and lift, which includes zero-flip problems and applies arm-specific pairwise retention. These populations need not coincide; truncation and unavailable required observations reduce retention.
Every estimate is shown with a pointwise 95\% problem-bootstrap interval. All StrategyQA directional swings have individual McNemar $p<.05$.

\begin{table}[htbp]
\caption{Confidence--doubt delegation swing ($S$), well-targeted flip share ($\mathrm{WT}$), random-selection reference ($\mathrm{WT}_{\mathrm{rand}}$), and targeting lift for every experiment, open continuation. $S$ and lift are in percentage points; WT and $\mathrm{WT}_{\mathrm{rand}}$ are percentages. H/F gives helpful/all flip events. Brackets are pointwise 95\% problem-bootstrap intervals. Appendix~\ref{app:results-table} describes the columns and retained populations.}
\label{tab:null-full}
\centering
\scriptsize
\resizebox{\linewidth}{!}{\begin{tabular}{llrrrrrr}
\toprule
Task & Model & $n_S/n_{\mathrm{WT}}$ & $S$ [CI] & H/F & WT [CI] & $\mathrm{WT}_{\mathrm{rand}}$ [CI] & Lift [CI] \\
\midrule
\multicolumn{8}{l}{\textbf{Open-weight models} ($n=1000$)} \\[1pt]
MuSiQue & Qwen3-4B & 966/876 & 14.0 [11.8, 16.3] & 78/134 & 58.2 [49.2, 67.2] & 39.4 [35.7, 43.1] & 18.8 [10.7, 27.1] \\
MuSiQue & Qwen3-8B & 944/921 & 22.4 [19.6, 25.1] & 128/301 & 42.5 [36.4, 48.8] & 40.6 [37.1, 44.1] & 1.9 [-3.1, 7.1] \\
MuSiQue & Qwen3-14B & 990/986 & 31.0 [28.0, 34.0] & 134/321 & 41.7 [36.4, 47.1] & 36.5 [33.5, 39.6] & 5.2 [0.9, 9.6] \\
MuSiQue & Qwen3-32B & 975/981 & 38.3 [35.1, 41.3] & 136/366 & 37.2 [32.1, 42.2] & 30.4 [27.4, 33.3] & 6.8 [3.0, 10.7] \\
MuSiQue & Gemma4-E2B & 1000/993 & 24.2 [21.5, 27.0] & 108/293 & 36.9 [31.1, 42.9] & 44.7 [41.1, 48.3] & -7.9 [-12.6, -3.0] \\
MuSiQue & Gemma4-E4B & 1000/1000 & 66.5 [63.5, 69.5] & 190/653 & 29.1 [25.6, 32.7] & 29.5 [26.4, 32.8] & -0.4 [-2.0, 1.1] \\
MuSiQue & Gemma4-31B & 957/953 & 1.3 [0.6, 2.0] & 5/15 & 33.3 [6.7, 63.6] & 39.5 [17.2, 65.8] & -6.2 [-16.2, 9.4] \\
MuSiQue & GLM-Z1-9B & 874/788 & 3.7 [2.4, 5.0] & 14/39 & 35.9 [20.5, 51.9] & 38.1 [34.5, 41.9] & -2.2 [-17.3, 13.5] \\
MuSiQue & GLM-Z1-32B & 800/767 & 5.6 [4.1, 7.2] & 12/44 & 27.3 [14.3, 41.2] & 30.5 [27.0, 34.0] & -3.3 [-15.9, 10.3] \\
StrategyQA & Qwen3-4B & 997/989 & 22.2 [19.5, 25.0] & 149/352 & 42.3 [36.4, 48.3] & 40.9 [36.5, 45.3] & 1.4 [-2.3, 5.3] \\
StrategyQA & Qwen3-8B & 961/992 & 19.0 [16.5, 21.6] & 90/228 & 39.5 [32.8, 46.4] & 35.7 [30.3, 41.1] & 3.8 [0.3, 7.3] \\
StrategyQA & Qwen3-14B & 995/998 & 33.9 [31.0, 36.8] & 189/384 & 49.2 [44.0, 54.6] & 46.3 [42.3, 50.5] & 2.9 [-0.3, 6.1] \\
StrategyQA & Qwen3-32B & 1000/1000 & 38.3 [35.3, 41.4] & 260/475 & 54.7 [49.8, 59.8] & 53.0 [48.8, 57.3] & 1.7 [-0.7, 4.1] \\
StrategyQA & Gemma4-E2B & 1000/999 & 7.3 [5.7, 9.0] & 41/80 & 51.2 [40.0, 62.5] & 45.4 [40.3, 50.4] & 5.9 [-4.3, 15.8] \\
StrategyQA & Gemma4-E4B & 1000/1000 & 25.8 [23.1, 28.5] & 78/255 & 30.6 [24.8, 36.5] & 30.0 [25.2, 34.8] & 0.6 [-2.6, 3.8] \\
StrategyQA & Gemma4-31B & 998/985 & 17.3 [14.9, 19.8] & 89/202 & 44.1 [36.7, 51.3] & 33.0 [28.1, 38.0] & 11.1 [5.7, 16.5] \\
StrategyQA & GLM-Z1-9B & 982/980 & 10.1 [8.2, 12.1] & 48/112 & 42.9 [33.3, 52.5] & 39.9 [33.7, 45.9] & 3.0 [-5.0, 10.7] \\
StrategyQA & GLM-Z1-32B & 987/990 & 7.2 [5.5, 8.9] & 40/88 & 45.5 [34.7, 56.2] & 43.5 [35.6, 51.2] & 2.0 [-5.6, 9.4] \\
\specialrule{1pt}{2pt}{3pt}
\multicolumn{8}{l}{\textbf{Provider-served models} ($n=250$)} \\[1pt]
MuSiQue & DeepSeek-V4-Pro & 250/250 & 66.8 [61.2, 72.4] & 48/167 & 28.7 [21.9, 35.8] & 24.1 [18.8, 29.6] & 4.6 [1.5, 7.8] \\
StrategyQA & DeepSeek-V4-Pro & 250/249 & 70.0 [64.4, 75.6] & 84/176 & 47.7 [40.4, 55.2] & 48.9 [42.1, 55.7] & -1.1 [-3.7, 1.5] \\
MuSiQue & MiniMax-M3 & 244/238 & 66.4 [60.2, 72.1] & 46/170 & 27.1 [20.3, 34.1] & 25.8 [19.9, 32.0] & 1.2 [-2.0, 4.5] \\
StrategyQA & MiniMax-M3 & 249/250 & 53.0 [46.6, 59.0] & 41/135 & 30.4 [22.6, 38.5] & 30.4 [23.8, 37.3] & -0.1 [-4.6, 4.3] \\%
\bottomrule
\end{tabular}}
\end{table}

\subsection{Provider-Served Models}
\label{app:frontier}

The provider-served extension uses the same two datasets, answer-confidence sentences, action parser, unaided competence labels, and deterministic post-processing as the open-weight study. DeepSeek-V4-Pro and MiniMax-M3 were run on $n=250$ sampled problems per task. The serving routes are DeepSeek-V4-Pro on CoreWeave (FP8) and MiniMax-M3 on Together (precision not reported). Their complete results appear in Table~\ref{tab:null-full} in Appendix~\ref{app:results-table}.

At $n=250$, both provider-served models respond strongly to the intervention but select problems little better than random: across the four experiments, $S$ ranges from $53.0$ to $70.0\pp$, $\mathrm{WT}$ from $27.1\%$ to $47.7\%$, and the targeting lift from $-1.1$ to $+4.6\pp$ (Table~\ref{tab:null-full}). In the MiniMax-M3 MuSiQue experiment, six with-tool bases contain no usable completed reasoning segment, leaving 244 splice continuations; unaided accuracy uses 241 completed self-answers.

Provider-side continuation semantics and deterministic replay are less directly auditable than open-weight decoding. These experiments are reported separately and are not pooled with the open-weight summaries.

\subsection{Flip composition}
\label{app:flip-composition}

\begin{figure*}[!ht]
\centering
\includegraphics[width=\linewidth]{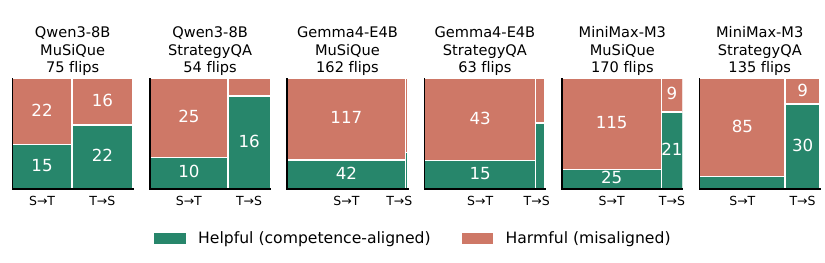}
\caption{Flip-event composition by direction (S$\rightarrow$T: \selftotool{}; T$\rightarrow$S: \tooltoself{}) and by alignment with unaided competence (helpful or harmful) for three illustrative models on both tasks, all at nominal $n=250$.}
\label{fig:flip-transitions}
\end{figure*}

Figure~\ref{fig:flip-transitions} decomposes the flip events behind WT for three illustrative models, Qwen3-8B, Gemma4-E4B, and the provider-served MiniMax-M3, on both tasks (all at nominal $n=250$). Positive and negative interventions are pooled within each task, including counterdirectional flips.
Each panel first splits the flips by direction, \selftotool{} (S$\rightarrow$T) and \tooltoself{} (T$\rightarrow$S), with widths proportional to event counts, and then splits each direction by height into competence-aligned (helpful) and misaligned (harmful) events. Because each panel has unit total area, the helpful area equals the pooled well-targeted share for that model and task, and the panel shows which direction contributes the helpful and harmful flips that make up that ratio.

\subsection{Problem-Sampling Uncertainty}
\label{app:uncertainty}

Table~\ref{tab:null-full} reports observed 95\% paired nonparametric bootstrap intervals for all 22 experiments: the 18 open-weight experiments at nominal $n=1000$ and the four provider-served experiments at $n=250$. Each of 20,000 replicates resamples problem IDs with replacement within a model--task experiment and recomputes $S$ or the ratio in Eq.~\ref{eq:well-targeted-share}; the latter therefore preserves zero-flip problems and the random flip denominator. These intervals quantify problem-sampling uncertainty under the fixed seed and decoding configuration, not uncertainty across seeds, temperatures, or serving stacks.

\subsection{Sampled-Decoding Sensitivity}
\label{app:stochastic}

The sensitivity pilot fixes the persisted $n=100$, problem-seed-0 sample and runs Qwen3-8B, Gemma4-E4B, and GLM-Z1-9B on both tasks at temperature 1.0 and top-$p=0.95$. Decode seeds 1--3 determine per-request streams through a hash of decode seed and dataset index. Positive and negative splice arms are therefore seed-paired, but base and splice generations are not draw-aligned because they consume randomness from different continuation boundaries. Greedy is reported separately rather than treated as a fourth seed.

The pooled estimator combines the three seed-specific problem contributions and resamples dataset-index clusters, retaining all observations of a problem across seeds in each of 20,000 bootstrap replicates. Of 7,200 planned generations, 7,190 are realized: five base/problem-seed instances lack an eligible completed reasoning boundary, removing ten positive/negative continuations across six splice experiments. No base row is missing.

\begin{table}[h]
\caption{Stochastic-decoding sensitivity at temperature 1.0 and top-$p=0.95$. $S_0$ is the greedy estimate; $S_1$--$S_3$ are decode-seed estimates. $S_{\mathrm{pool}}$ pools seed-paired problem contributions, and brackets give a 95\% problem-clustered bootstrap interval. All values are percentage points.}
\label{tab:stochastic-robustness}
\centering
\scriptsize
\setlength{\tabcolsep}{3.5pt}
\begin{tabular}{llrrrrr}
\toprule
Model & Task & $S_0$ & $S_1$ & $S_2$ & $S_3$ & $S_{\mathrm{pool}}$ [95\% CI] \\
\midrule
Qwen3-8B & StrategyQA & 24.2 & 19.4 & 14.0 & 21.6 & 18.2 [13.1, 23.7] \\
Qwen3-8B & MuSiQue & 21.4 & 17.5 & 11.2 & 15.5 & 14.7 [9.1, 20.6] \\
Gemma4-E4B & StrategyQA & 24.2 & 18.6 & 21.3 & 22.7 & 20.8 [15.5, 26.3] \\
Gemma4-E4B & MuSiQue & 70.0 & 40.4 & 45.1 & 43.5 & 43.0 [36.2, 49.8] \\
GLM-Z1-9B & StrategyQA & 8.1 & 14.1 & 11.1 & 15.2 & 13.5 [9.1, 18.5] \\
GLM-Z1-9B & MuSiQue & 3.4 & 5.0 & 6.0 & 1.0 & 4.0 [1.7, 6.7] \\
\bottomrule
\end{tabular}
\end{table}
 \begin{table}[h]
\caption{Unaided accuracy and well-targeted share under stochastic decoding (temperature 1.0, top-$p=0.95$, nominal $n=100$). Subscript 0 is the greedy run and 1--3 are decode seeds; each seed relabels competence from its own sampled no-tool answers. $\mathrm{WT}_{\mathrm{pool}}$ pools seed-paired flip contributions, with a 95\% problem-clustered bootstrap interval. All values are percent.}
\label{tab:stochastic-accuracy-wt}
\centering
\scriptsize
\setlength{\tabcolsep}{3pt}
\begin{tabular}{ll rrrr rrrr r}
\toprule
&& \multicolumn{4}{c}{Unaided accuracy} & \multicolumn{5}{c}{WT} \\
\cmidrule(lr){3-6} \cmidrule(lr){7-11}
Model & Task & $0$ & $1$ & $2$ & $3$ & $0$ & $1$ & $2$ & $3$ & pool [95\% CI] \\
\midrule
Qwen3-8B & StrategyQA & 72.0 & 74.0 & 75.0 & 75.0 & 52.2 & 32.1 & 43.5 & 29.6 & 34.6 [22.8, 47.9] \\
Qwen3-8B & MuSiQue & 64.9 & 64.0 & 66.0 & 62.0 & 51.5 & 58.5 & 56.2 & 27.0 & 47.3 [35.9, 58.4] \\
Gemma4-E4B & StrategyQA & 70.0 & 68.0 & 67.0 & 63.0 & 36.0 & 66.7 & 63.9 & 40.5 & 56.0 [44.9, 67.0] \\
Gemma4-E4B & MuSiQue & 62.0 & 62.0 & 67.0 & 65.0 & 25.0 & 21.6 & 34.0 & 35.0 & 29.7 [20.4, 39.8] \\
GLM-Z1-9B & StrategyQA & 72.7 & 71.4 & 68.0 & 70.7 & 50.0 & 68.8 & 26.3 & 60.0 & 50.0 [36.1, 64.1] \\
GLM-Z1-9B & MuSiQue & 62.5 & 57.0 & 60.0 & 57.0 & 66.7 & 60.0 & 40.0 & 60.0 & 50.0 [25.0, 75.0] \\
\bottomrule
\end{tabular}
\end{table}
 
All six experiments meet the specified directional criterion, and every pooled interval excludes zero. The stochastic estimates therefore support the qualitative claim that doubt raises delegation relative to confidence.
Targeting is less stable across seeds than sensitivity: seed-specific WT estimates cross the descriptive 50\% majority threshold in several experiments, and only Gemma4-E4B on MuSiQue preserves the greedy majority side in the pooled estimate and at least two seeds. Because three seeds were selected for a directional stress test rather than variance-component estimation, we do not report a between-seed standard error. Table~\ref{tab:stochastic-accuracy-wt} separates the two: unaided accuracy varies by at most $7\pp$ across the greedy run and the three seeds, whereas WT varies by $13$--$43\pp$ within an experiment. Competence labels are therefore stable under sampling, and the instability lies in which problems flip, which at $n=100$ per seed also reflects small flip counts.

\subsection{Delegation Rates Across the Base and Splice Arms}
\label{app:placebo}

Table~\ref{tab:placebo-arms} reports the delegation rate under every condition that shares the same reasoning prefix: the uninterrupted with-tool base $B$, the sentence-less splice $\none$, the neutral sentence $\nudgeneutral$, and the two primary arms $\nudgepos$ and $\nudgeneg$. Reading across a row separates the parts of the intervention. Comparing $\none$ with $B$ isolates the cut-and-regeneration operation; comparing $\nudgeneutral$ with $\none$ isolates inserting a first-person sentence that carries no confidence content; the primary arms add confidence or doubt content on top. All five arms are pairwise complete on a single retained domain per model--task experiment at nominal $n=1000$, so each row has one denominator $n$ and its columns are directly comparable. Neither control enters $S$, $\mathrm{WT}$, or $L$.

\begin{table}[h]
\caption{Delegation rate $P(A=\tool)$, in percent, for the uninterrupted with-tool base $B$ and the four splice arms: sentence-less ($\none$), neutral ($\nudgeneutral$), confidence ($\nudgepos$), and doubt ($\nudgeneg$). Nominal $n=1000$; each row's arms share one pairwise-complete denominator $n$.}
\label{tab:placebo-arms}
\centering
\scriptsize
\begin{tabular}{llrrrrrr}
\toprule
Task & Model & $n$ & $B$ & $\none$ & $\nudgeneutral$ & $\nudgepos$ & $\nudgeneg$ \\
\midrule
MuSiQue & Qwen3-4B & 955 & 10.7 & 9.8 & 9.3 & 8.5 & 22.3 \\
 & Qwen3-8B & 930 & 36.0 & 35.4 & 27.4 & 27.7 & 50.1 \\
 & Qwen3-14B & 990 & 7.5 & 6.3 & 6.6 & 4.7 & 35.8 \\
 & Qwen3-32B & 970 & 10.1 & 7.8 & 8.2 & 6.9 & 45.2 \\
 & Gemma4-E2B & 1000 & 10.0 & 10.0 & 5.6 & 4.5 & 28.7 \\
 & Gemma4-E4B & 1000 & 14.8 & 14.7 & 15.3 & 13.4 & 79.9 \\
 & Gemma4-31B & 956 & 0.7 & 0.7 & 0.4 & 0.3 & 1.6 \\
 & GLM-Z1-9B & 859 & 18.7 & 18.9 & 18.9 & 18.2 & 21.8 \\
 & GLM-Z1-32B & 776 & 10.6 & 10.6 & 10.7 & 10.3 & 15.6 \\
\midrule
StrategyQA & Qwen3-4B & 997 & 73.7 & 75.3 & 73.0 & 67.1 & 89.3 \\
 & Qwen3-8B & 953 & 80.7 & 82.6 & 80.0 & 76.4 & 95.5 \\
 & Qwen3-14B & 994 & 76.6 & 76.6 & 74.7 & 59.8 & 93.7 \\
 & Qwen3-32B & 977 & 80.5 & 76.9 & 45.4 & 57.0 & 95.9 \\
 & Gemma4-E2B & 1000 & 79.1 & 79.1 & 77.9 & 76.1 & 83.4 \\
 & Gemma4-E4B & 1000 & 65.7 & 65.4 & 65.9 & 64.0 & 89.8 \\
 & Gemma4-31B & 998 & 32.5 & 32.5 & 32.3 & 26.6 & 43.9 \\
 & GLM-Z1-9B & 981 & 82.6 & 84.4 & 79.7 & 79.2 & 89.3 \\
 & GLM-Z1-32B & 984 & 88.6 & 88.9 & 87.5 & 85.6 & 92.8 \\
\bottomrule
\end{tabular}
\end{table}

Table~\ref{tab:placebo-arms} reports both controls against the uninterrupted base. The
sentence-less splice is close to inert: the median absolute difference between $\none$ and $A^B$
is $0.3\pp$ and the largest is $3.6\pp$, consistent with a cut-and-regeneration that is
deterministic under greedy decoding. The neutral sentence is usually but not always inert. Its
median shift from $\none$ is $-0.9\pp$, yet it moves delegation by at least $3\pp$ in four of the
eighteen experiments, and for Qwen3-32B on StrategyQA it lowers delegation by $31.4\pp$, further
than the confidence sentence does. In those four cells $\nudgeneutral$ behaves like $\nudgepos$
rather than like a placebo, which is why $S$ is defined as a confidence--doubt contrast rather
than against the neutral arm.

\subsection{Sealed-Continuation Sanity Check}
\label{app:closed}

At the original nominal $n=1000$ scope, the closed continuation removes the model's opportunity to reason after reading the inserted statement. Across models, the difference between closed and open swing ranges from $-20.8\pp$ to $+53.7\pp$. Sealing amplifies the median swing on MuSiQue ($22.4\pp$ open versus $30.7\pp$ closed) but not on StrategyQA ($19.0\pp$ open versus $13.4\pp$ closed). The open-versus-sealed difference varies in sign across models; this contrast does not separate the effect of deliberation from that of the changed boundary. Because sealing also changes the continuation boundary, we treat it as a sanity check rather than pooling it into the primary result.

\subsection{Natural Delegation and the Oracle Control}
\label{app:natural-oracle}

The uninterrupted tool-aware runs provide a baseline for how well models route without an inserted signal. The natural targeting gap $\Gamma$ is positive in 17 of the 18 open-weight model--task experiments, ranging from $-2.5\pp$ to $+26.6\pp$, median +5.6 pp. Thus models often delegate more on problems they fail unaided, but the strength of that association varies widely. Neither higher unaided accuracy nor larger parameter count produces a monotone delegation pattern.

Replacing the scoped tool with a free perfect oracle does not eliminate unnecessary self-reliance. Across the open-weight experiments, models still answer directly on $18.9\%$--$96.2\%$ of retained problems despite access to a costless, guaranteed-correct alternative. Among those retained self-answers, $11.1\%$--$45.1\%$ occur on problems the model failed unaided. The spread is driven mainly by task: under the oracle, models answer directly on $64$--$96\%$ of MuSiQue problems but on $19$--$44\%$ of StrategyQA problems, Gemma4-31B ($79\%$) excepted. The oracle also does not raise delegation overall ($\Phi$ has median $-1.5\pp$ and is positive in 8 of 18 experiments; Table~\ref{tab:oracle}), but it sharpens its targeting: on the problems both bases retain, $\Gamma$ grows in 14 of 18 experiments (median $5.4$ to $10.2\pp$) and is positive in all 18. Removing search uncertainty and request-formulation effort therefore does not fully explain non-delegation. Some models continue to answer precisely on cases for which their observed unaided behavior indicates that help is needed. Appendix~\ref{app:retained-self-correctness} reports correctness transitions among the retained self-answers.

\subsection{Retained Self-Answer Correctness}
\label{app:retained-self-correctness}

This supporting correctness analysis retains nominal $n=1000$. The no-tool and with-tool bases are separate greedy generations on the same sampled problems. Table~\ref{tab:retained-self-correctness} conditions on rows where the no-tool answer is gradeable and the with-tool base commits to \self. Consequently, $D_{\mathrm{self}}$ and $R_{\mathrm{self}}$ measure correctness transitions among retained self-answers, not the value of delegation. They also do not assert that the answer text or reasoning was held fixed: adding the tool definition changes the prompt and can change the entire generated continuation even when the final action remains \self.

\begingroup
\let\scopeCaption\caption
\renewcommand{\caption}[1]{\scopeCaption{Supporting analysis at nominal $n=1000$. #1}}
\begin{table}[h]
\caption{Correctness transitions when the tool-aware base still answers for itself. Rates are percentages. $D_{\mathrm{self}}$ is correct-to-incorrect degradation and $R_{\mathrm{self}}$ is incorrect-to-correct recovery; parentheses give numerator/eligible retained-self denominator.}
\label{tab:retained-self-correctness}
\centering
\scriptsize
\begin{tabular}{llrr}
\toprule
Task & Model & $D_{\mathrm{self}}$ (count) & $R_{\mathrm{self}}$ (count) \\
\midrule
MuSiQue & Qwen3-4B & 10.0 (49/489) & 12.2 (38/312) \\
 & Qwen3-8B & 8.0 (33/411) & 11.9 (21/177) \\
 & Qwen3-14B & 9.5 (56/587) & 12.3 (39/316) \\
 & Qwen3-32B & 5.6 (34/607) & 12.7 (32/251) \\
 & Gemma4-E2B & 21.0 (99/471) & 11.1 (47/423) \\
 & Gemma4-E4B & 4.7 (25/527) & 24.8 (54/218) \\
 & Gemma4-31B & 3.0 (24/787) & 11.9 (17/143) \\
 & GLM-Z1-9B & 14.3 (61/428) & 10.4 (26/251) \\
 & GLM-Z1-32B & 6.1 (30/495) & 13.1 (29/222) \\
\midrule
StrategyQA & Qwen3-4B & 9.0 (18/199) & 24.6 (15/61) \\
 & Qwen3-8B & 6.2 (10/160) & 21.1 (8/38) \\
 & Qwen3-14B & 5.3 (10/189) & 11.4 (5/44) \\
 & Qwen3-32B & 9.8 (16/164) & 25.9 (7/27) \\
 & Gemma4-E2B & 27.1 (35/129) & 16.9 (11/65) \\
 & Gemma4-E4B & 27.7 (65/235) & 16.0 (13/81) \\
 & Gemma4-31B & 2.9 (17/595) & 11.8 (9/76) \\
 & GLM-Z1-9B & 8.1 (11/135) & 21.1 (8/38) \\
 & GLM-Z1-32B & 2.1 (2/95) & 23.5 (4/17) \\
\midrule
All self-deployed experiments & Pooled & 8.9 (595/6703) & 13.9 (383/2760) \\
\bottomrule
\end{tabular}
\end{table}
 \endgroup

\subsection{Supporting Control Details}
\label{app:controls}

\paragraph{Voice ($n=1000$).}
Table~\ref{tab:voice} shows the first-person and expert endpoints; the second-person and reviewer arms are retained in the local per-experiment reports. Expert attribution exceeds first person only for Qwen3-14B and Qwen3-32B on StrategyQA, plus the tied/near-tied GLM-Z1-9B MuSiQue experiment under the report's comparison rule; in the remaining experiments self-attribution is at least as strong.

\begin{table}[h]
\caption{Supporting voice control at nominal $n=1000$: open-continuation swing by voice endpoint, in percentage points.}
\label{tab:voice}
\centering
\scriptsize
\resizebox{\linewidth}{!}{\begin{tabular}{llrr@{\qquad}llrr}
\toprule
Task & Model & First person & Expert & Task & Model & First person & Expert \\
\midrule
MuSiQue & Qwen3-4B & 14.0 & 2.7 & StrategyQA & Qwen3-4B & 22.2 & 15.6 \\
 & Qwen3-8B & 22.4 & 12.8 &  & Qwen3-8B & 19.0 & 15.7 \\
 & Qwen3-14B & 31.0 & 11.4 &  & Qwen3-14B & 33.9 & 47.9 \\
 & Qwen3-32B & 38.3 & 13.8 &  & Qwen3-32B & 38.3 & 39.2 \\
 & Gemma4-E2B & 24.2 & 10.3 &  & Gemma4-E2B & 7.3 & 2.4 \\
 & Gemma4-E4B & 66.5 & 37.2 &  & Gemma4-E4B & 25.8 & 11.0 \\
 & Gemma4-31B & 1.3 & 0.1 &  & Gemma4-31B & 17.3 & 10.2 \\
 & GLM-Z1-9B & 3.7 & 3.5 &  & GLM-Z1-9B & 10.1 & 7.8 \\
 & GLM-Z1-32B & 5.6 & 1.6 &  & GLM-Z1-32B & 7.2 & 6.1 \\
\bottomrule
\end{tabular}}

\end{table}

\paragraph{Free oracle ($n=1000$).}
Let $\Phi=P(\tool\mid\text{oracle})-P(\tool\mid\text{scoped})$ measure the delegation bought by eliminating request formulation. Table~\ref{tab:oracle} reports $\Phi$, residual self-reliance $\rho$, and conditional overconfidence $\omega$.

\begin{table}[h]
\caption{Supporting oracle control at nominal $n=1000$: decomposition in percent/percentage points. Negative $\Phi$ means the model delegates less to the generic free oracle than to the scoped tool.}
\label{tab:oracle}
\centering
\scriptsize
\resizebox{\linewidth}{!}{\begin{tabular}{llrrr@{\qquad}llrrr}
\toprule
Task & Model & $\Phi$ & $\rho$ & $\omega$ & Task & Model & $\Phi$ & $\rho$ & $\omega$ \\
\midrule
MuSiQue & Qwen3-4B & -1.2 & 90.0 & 39.6 & StrategyQA & Qwen3-4B & -17.9 & 44.1 & 24.5 \\
 & Qwen3-8B & -1.8 & 64.3 & 29.0 &  & Qwen3-8B & -2.8 & 22.7 & 18.9 \\
 & Qwen3-14B & +0.3 & 92.2 & 34.1 &  & Qwen3-14B & -20.2 & 43.5 & 17.5 \\
 & Qwen3-32B & +2.0 & 87.9 & 29.4 &  & Qwen3-32B & +0.2 & 18.9 & 11.1 \\
 & Gemma4-E2B & +6.5 & 83.5 & 45.1 &  & Gemma4-E2B & -12.9 & 33.8 & 34.2 \\
 & Gemma4-E4B & +7.7 & 77.5 & 27.7 &  & Gemma4-E4B & -7.5 & 41.8 & 20.5 \\
 & Gemma4-31B & +3.0 & 96.2 & 15.0 &  & Gemma4-31B & -11.3 & 78.8 & 12.4 \\
 & GLM-Z1-9B & +0.5 & 80.1 & 37.6 &  & GLM-Z1-9B & -12.0 & 29.5 & 17.2 \\
 & GLM-Z1-32B & +4.3 & 84.4 & 31.9 &  & GLM-Z1-32B & -18.7 & 30.0 & 13.4 \\
\bottomrule
\end{tabular}}

\end{table}

The mostly non-positive StrategyQA values show that a generic oracle description is not simply a more attractive tool. This control isolates willingness to defer under a guaranteed-correct option; it does not imply that the scoped tool is valueless or that a real remote model is free and perfect.

\end{document}